\pdfoutput=1

\documentclass[11pt]{article}

\usepackage[preprint]{acl}

\makeatletter
\ifacl@anonymize
\gdef\isanonymous{1}
\fi

\usepackage{times}
\usepackage{latexsym}
\usepackage{footnote}

\usepackage[T1]{fontenc}

\usepackage[utf8]{inputenc}

\usepackage{microtype}

\usepackage{inconsolata}

\usepackage{graphicx}

\usepackage[british]{babel}

\usepackage{subfigure}
\usepackage{booktabs} %
\usepackage[acronym,shortcuts]{glossaries}
\usepackage{glossaries-prefix}
\usepackage{xspace}
\usepackage{multirow}
\usepackage{makecell}
\usepackage{adjustbox}
\usepackage{setspace}
\usepackage[most]{tcolorbox}
\usepackage{tabularx}
\usepackage[inline]{enumitem}
\usepackage{fontawesome5}
\usepackage{calc}
\usepackage{bibentry}
\usepackage[dvipsnames,table]{xcolor}
\usepackage{listings}
\usepackage{hyperref}
\usepackage{longtable}
\nobibliography* %

\glsdisablehyper

\usepackage{amsmath}
\usepackage{amssymb}
\usepackage{mathtools}
\usepackage{amsthm}

\usepackage{amsmath}      %
\usepackage{amssymb}      %
\usepackage{algorithm}    %
\usepackage{algpseudocode}%

\usepackage[capitalize]{cleveref}

\usepackage[textsize=tiny]{todonotes}
\usepackage{pifont}
\newcommand{\cmark}{\textcolor{ForestGreen}{\ding{51}}\xspace}
\newcommand{\xmark}{\textcolor{Red}{\ding{55}}\xspace}

\newcommand{\errorinstallfailed}{\xmark}

\makeatletter
\DeclareRobustCommand\onedot{\futurelet\@let@token\@onedot}
\def\@onedot{\ifx\@let@token.\else.\null\fi\xspace}

\def\eg{\emph{e.g}\onedot} 
\def\ie{\emph{i.e}\onedot} 
\def\cf{\emph{c.f}\onedot} 
\def\etc{\emph{etc}\onedot}

\makeatother

\newacronym[prefixfirst={a\ },prefix={an\ }]
  {llm}{LLM}{large language model}
\newacronym[prefixfirst={an\ },prefix={an\ }]
  {ai}{AI}{artificial intelligence}
\newacronym[prefixfirst={a\ },prefix={an\ }]
  {http}{HTTP}{HyperText Transfer Protocol}
\newacronym[prefixfirst={a\ },prefix={a\ }]
  {wsi}{WSI}{whole slide image}
\newacronym[prefixfirst={an\ },prefix={an\ }]
  {api}{API}{application programming interface}
\newacronym[prefixfirst={a\ },prefix={a\ }]
  {gpu}{GPU}{graphics processing unit}
\newacronym[prefixfirst={an\ },prefix={an\ }]
  {os}{OS}{operating system}

\glsunset{http}
\glsunset{api}
\glsunset{gpu}

\newlength{\iconTotalHeight}
\newlength{\iconDepth}
\newlength{\iconRaise}

\NewDocumentCommand{\icon}{m}{%
  \settototalheight{\iconTotalHeight}{\strut ABCDEFGHIJKLMNOPQRSTUVWXYZabcdefghijklmnopqrstuvwxyz}%
  \settodepth{\iconDepth}{\strut q}%
  \setlength{\iconRaise}{\dimexpr -\iconDepth + 0.08\iconTotalHeight\relax}%
  \raisebox{\iconRaise}{%
    \includegraphics[height=0.9\iconTotalHeight]{#1}%
  }%
}

\definecolor{llmflow}{HTML}{60B258}
\definecolor{controlflow}{HTML}{6C6B6B}
\definecolor{environmentflow}{HTML}{1D63ED}
\definecolor{agentflow}{HTML}{F59E0B}

\makeatletter
\AddToHook{cmd/appendix/before}{%
  \def\cref@section@alias{appendix}%
  \def\cref@section@alias@plural{appendices}%
  \def\cref@subsection@alias{appendix}%
  \def\cref@subsection@alias@plural{appendices}%
}
\makeatother

\newlength{\oldfboxsep}
\newcommand{\coloredbox}[2]{%
  \begingroup%
  \setlength{\oldfboxsep}{\fboxsep}%
  \setlength\fboxsep{0pt}%
  \colorbox{#1}{\strut \hspace{.5pt} #2 \hspace{.5pt}}%
  \setlength\fboxsep{\oldfboxsep}%
  \endgroup%
}

\usepackage[normalem]{ulem}

\newcommand{\ours}{\mbox{\textsc{ToolMaker}}\xspace}
\newcommand{\Ours}{\ours}
\newcommand{\ourbenchmark}{\mbox{\textsc{TM-Bench}}\xspace}
\newcommand{\Ourbenchmark}{\ourbenchmark}

\title{LLM Agents Making Agent Tools}

\author{
  \textbf{Georg W\"{o}lflein\textsuperscript{\normalfont 1,2,3,\textdagger}}\quad
  \textbf{Dyke Ferber\textsuperscript{\normalfont 3,4,\textdaggerdbl}}\quad
  \textbf{Daniel Truhn\textsuperscript{\normalfont 5}}\\
  \textbf{Ognjen Arandjelovi\'{c}\textsuperscript{\normalfont 2}}\quad
  \textbf{Jakob N.~Kather\textsuperscript{\normalfont 1,4,6}}
  \vspace{0.7em}
  \\
  \textsuperscript{1}EKFZ for Digital Health, TU Dresden
  \quad
  \textsuperscript{2}University of St Andrews
  \\
  \textsuperscript{3}Synagen AI
  \quad
  \textsuperscript{4}NCT, Heidelberg University Hospital
  \\
  \textsuperscript{5}University Hospital Aachen
  \quad
  \textsuperscript{6}University Hospital Dresden
  \\
  \small{
    \textbf{Correspondence:} \href{mailto:georg@woelflein.de}{georg@woelflein.de}
  }
}

\colorlet{cellgreen}{green!20}
\colorlet{cellyellow}{yellow!20}
\colorlet{cellred}{red!20}

\usepackage{titlesec}
\titlespacing{\paragraph}{%
  0pt}{%
  0.2\baselineskip}{%
  1em}%

\usepackage{setspace}
\begin{document}

\setlength{\abovedisplayskip}{3pt}%
\setlength{\belowdisplayskip}{3pt}%
\setlength{\abovedisplayshortskip}{3pt}%
\setlength{\belowdisplayshortskip}{3pt}%

\maketitle

\renewcommand{\thefootnote}{\fnsymbol{footnote}}
\setcounter{footnote}{0}
\footnotetext[2]{ Work done while at EKFZ for Digital Health, TU Dresden and University of St Andrews.
\textsuperscript{\textdaggerdbl}Work done while at EKFZ for Digital Health, TU Dresden and NCT Heidelberg.}
\renewcommand{\thefootnote}{\arabic{footnote}}

\begin{abstract}
  Tool use has turned \glspl{llm} into powerful agents that can perform complex multi-step tasks by dynamically utilising external software components. 
  However, these tools must be implemented in advance by human developers, hindering the applicability of \gls{llm} agents in domains demanding large numbers of highly specialised tools, like in life sciences and medicine.
  Motivated by the growing trend of scientific studies accompanied by public code repositories, we propose \ours, an agentic framework that autonomously transforms papers with code into \gls{llm}-compatible tools.
  Given a GitHub URL and short task description, \ours autonomously installs dependencies and generates code to perform the task, using a closed-loop self-correction mechanism for debugging. 
  To evaluate our approach, we introduce a benchmark comprising 15 complex computational tasks spanning various domains with over 100 unit tests to assess correctness and robustness.
  Our method correctly implements 80\% of the tasks, substantially outperforming current state-of-the-art software engineering agents.
  \Ours therefore is a step towards fully autonomous agent-based scientific workflows\footnote{Our code and benchmark are publicly available at \mbox{\url{https://github.com/KatherLab/ToolMaker}}.}.
\end{abstract}

\begin{figure}[t]
  \includegraphics[width=\linewidth]{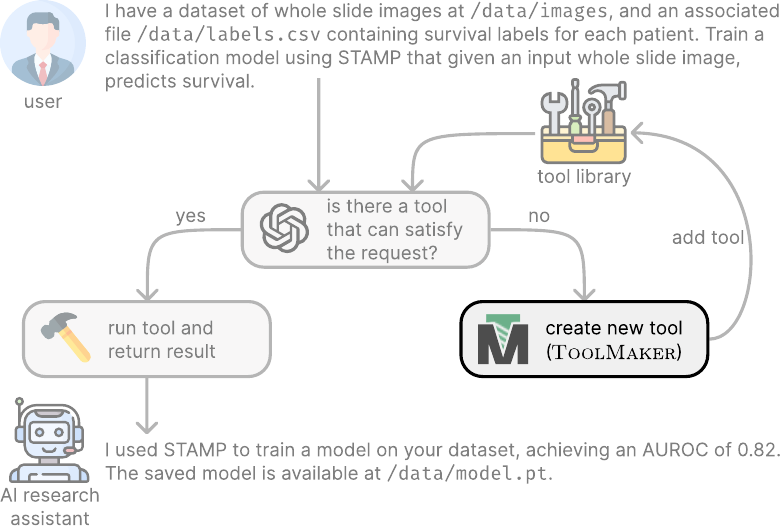}
  \caption{We envision a future where agents posess dynamic toolsets that can be expanded at runtime. \emph{Tool creation}, studied here, is a crucial step towards this goal.}
  \label{fig:ai_research_assistant}
\end{figure}

\section{Introduction}

Scientific discovery is the foundation for innovation and progress.
Traditionally, the underlying research processes that guarantee progress have been entirely reliant on human expertise, involving the formulation of ideas and hypotheses, the collection of information and analysis of data, the planning and execution of experiments, and iterative refinement to arrive at a solution. 
With the recent development of autonomous agents that employ \glspl{llm} to perform tasks through multi-step reasoning and planning, and by utilising tools (external pieces of software that the model can execute), we are at the cusp of a paradigm shift where \gls{ai} can assist throughout entire research projects as a \emph{virtual scientist} (\cref{fig:ai_research_assistant}), rather than being limited to addressing narrowly and \emph{a priori} defined problems.

Although \gls{llm} agents have shown success for \emph{specific} tasks in domains such as software engineering~\cite{wang2024openhands,yang2024sweagent}, healthcare~\cite{ferber2024autonomous,kim2024mdagents}, law~\cite{li2024legalagentbench}, and scientific research~\cite{swanson2024virtuallab,gao2024empowering,schmidgall2025agentlaboratory}, they struggle to generalise to broader classes of tasks. This limitation arises from their reliance on tools that must be explicitly designed, implemented, and integrated by human developers -- often requiring extensive technical expertise -- before deployment~\cite{ferber2024autonomous,jimenez2024swebench}. While \gls{ai} assistants can support this process, current systems still depend heavily on manual intervention to ensure compatibility and functionality.

To address this, some agentic frameworks have been designed that autonomously craft their own tools~\cite{cai2024toolmakers,yuan2024craft,qian2023creator}.
However, because these methods build each tool from scratch, they inevitably produce simple, narrowly scoped tools tailored to single-dimensional problems -- an approach ill-suited to the complexity of real-world research problems.

In fact, in critical fields such as healthcare, data necessary to build tools from scratch is often inaccessible due to privacy restrictions, preventing agents from using it to build their own solutions. 
Moreover, the complexity of modern scientific tools has increased substantially in terms of computational requirements, data demands, and amount of code involved. 
Lastly, deploying tools in high-stakes applications demands rigorous validation, testing, and quality assurance -- standards that current agent systems cannot realistically meet if required to develop such tools entirely from scratch.

Encouragingly, a growing emphasis on reproducibility within the scientific community has led to an increase in publicly released code accompanying research papers~\cite{zhou2024what}. 
Consequently, a vast array of potential tools now exist as standalone solutions. 
However, many researchers in fields like healthcare, biology, drug development, R\&D are unable to effectively use them due to the technical skills required for their deployment.

\begin{figure}[t]
  \includegraphics[width=\linewidth]{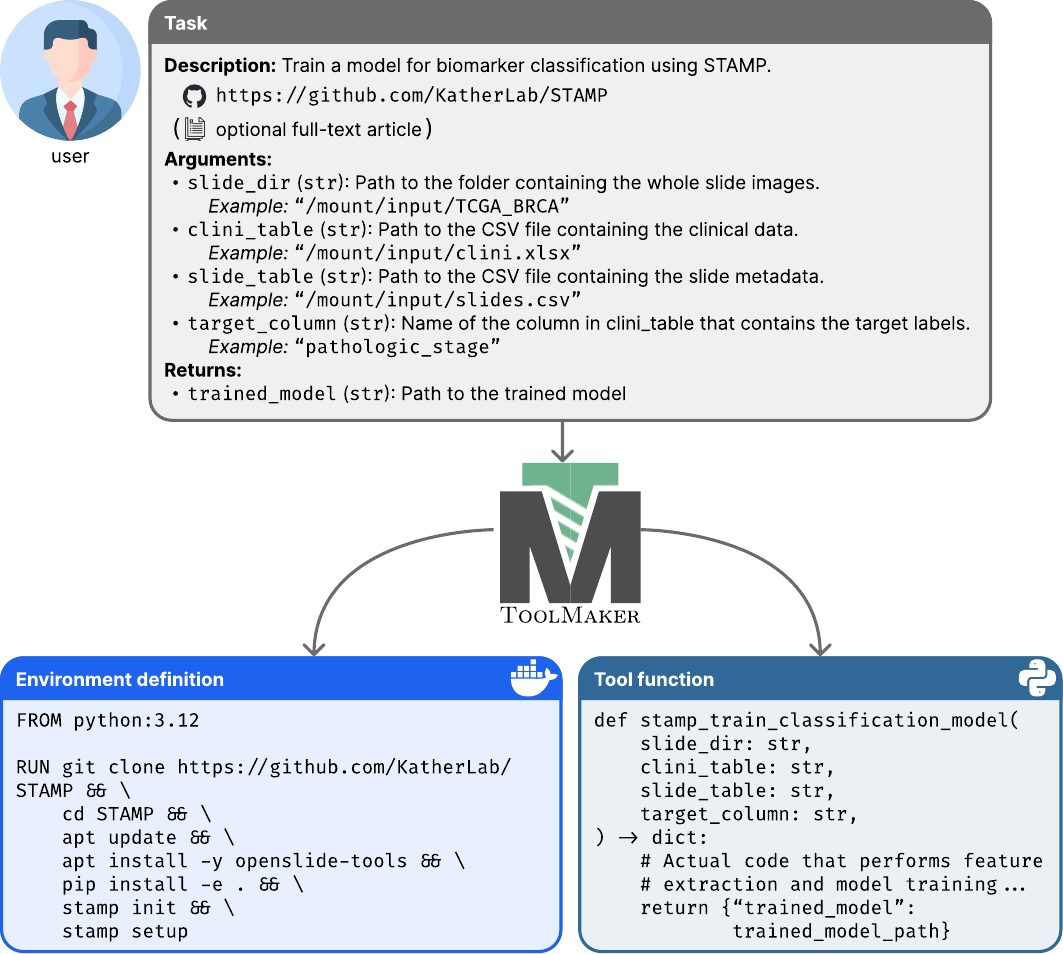}
  \caption{Given a task description, a scientific paper, a link to the associated code repository, and an example of the tool invocation, \ours creates (i) a Docker container in which the tool can be executed, (ii) a Python function that performs the task.}
  \label{fig:tool_creation}
\end{figure}

Instead of building tools entirely from scratch, we ask the following question: \emph{Can \gls{llm} agents autonomously download, integrate, and execute complex, existing tools to empower researchers with minimal technical expertise in the future?}
Towards this goal, we propose \ours, an agentic framework that autonomously generates LLM-compatible tools from scientific papers and their associated code repositories, bypassing the need for human intermediaries to manually set up, install, and adapt them to fit the requirements of their applications. Given a task description, a scientific paper, and its associated code repository, \ours generates an executable tool that enables \glspl{llm} to perform the task (see \cref{fig:tool_creation}).

To evaluate \ours, we introduce \ourbenchmark, a benchmark comprising 15 diverse tasks across various medical disciplines (pathology, radiology, omics), as well as non-medical fields, \eg \glspl{llm} and 3D vision. Unlike existing benchmarks~\cite{jimenez2024swebench,zhuo2024bigcodebench,jain2024livecodebench} which assume pre-installed dependencies for function implementation, \ours operates in a fully open-ended environment. Tasks in our benchmark encompass the entire workflow: downloading resources, managing and resolving dependency issues, reading through large codebases, and implementing, testing, and debugging code. 
\Ourbenchmark includes over 100 unit tests to objectively assess the generated tools' correctness.

\section{Related work}

\paragraph{Agents}
In addition to demonstrating impressive capabilities in generating human-like text, \glspl{llm} such as ChatGPT~\cite{ouyang2022training}, Claude~\cite{anthropic_claude}, Gemini~\cite{google2024gemini} and Llama~\cite{meta2024llama3}, on their own, have shown strong potential in question answering and reasoning on problems in natural science related fields, like math~\cite{shao2024deepseekmathpushinglimitsmathematical}, chemistry~\cite{bran2024augmenting} and healthcare~\cite{singhal2023llms}. However, \glspl{llm} often struggle solving more complex problems directly, especially in situations that require intermediate results from multiple steps~\cite{valmeekam2023planningabilitieslargelanguage}. 
To address this, \gls{llm} agents have been developed which enhance an \gls{llm}'s capabilities by integrating external tools~\cite{schick2023toolformer}.

In software engineering, a number of agentic and workflow-based systems have been proposed for solving GitHub issues~\cite{wang2024openhands,yang2024sweagent,xia2024agentless}, as well as developing entire software projects~\cite{qian2024chatdev,nguyen2024agilecoder,hong2024metagpt}.
Among these, OpenHands~\cite{wang2024openhands} achieves state-of-the-art performance on SWE-Bench~\cite{jimenez2024swebench}, a benchmark for solving GitHub issues. 

Medical \gls{llm} agents have been developed for clinical decision-making and diagnostics, such as building risk calculators from publications~\cite{jin2024agentmd}, oncology agents that consult guidelines and imaging tools~\cite{ferber2024autonomous}, and multi-agent systems that enable collaboration across clinicians, patients, and hospitals~\cite{kim2024mdagents,li2025agenthospital}.
Beyond clinical use, agents have been proposed for bioinformatics tasks like data extraction, pipeline execution, and hypothesis testing~\cite{ding2024automatingexploratoryproteomicsresearch,xin2024bioinformaticsagent}, even automating entire scientific projects, including literature reviews, experiment design, and manuscript writing~\cite{lu2024aiscientistfullyautomated,schmidgall2025agentlaboratory}.

Nonetheless, regardless of domain, agentic systems remain constrained by the tools at their disposal. 
For example, when tasked to solve a pathology image classification problem, the AIDE machine learning agent~\cite{weco2025introducing} trains a standard convolutional net (\cf Fig.~2 in \citet{chan2024mlebench}).
By contrast, a domain expert would instead employ pathology foundation models, as these have been designed specifically for this type of problem~\cite{chen2024uni,zimmermann2024virchow2,wolflein2023good}.
Thus, AIDE lacks the necessary tools to solve the task effectively.

\paragraph{Tool creation}
To address this, we consider the problem of \emph{tool creation} -- enabling \glspl{llm} to create their own tools, to dynamically expand their capabilities at runtime.
Tool creation is not to be confused with \emph{tool learning}, \ie teaching \glspl{llm} to utilise appropriate, human-crafted tools more effectively which has been extensively studied in recent years~\cite{qin2024toollearning,schick2023toolformer}.
Previous work on tool creation~\cite{cai2024toolmakers,yuan2024craft, qian2023creator} is limited to crafting very simple tools because (i) they are crafted from scratch, and (ii) these systems cannot interact with the \gls{os} by running bash commands, reading/writing files, \etc (see \cref{tab:tool-learning}). Our approach addresses both of these limitations.
\begin{table}[h]
  \adjustbox{max width=\linewidth}{
    \begin{tabular}{lcccc}
      \toprule
      \textbf{Method} & \makecell{\textbf{Error}\\\textbf{handling}} & \makecell{\textbf{OS}\\\textbf{interaction}} & \makecell{\textbf{Complex}\\\textbf{tasks}}                                           \\
      \midrule
      CRAFT~\cite{yuan2024craft} & \xmark & \xmark & \xmark \\
      CREATOR~\cite{qian2023creator} & \cmark & \xmark & \xmark \\
      LATM~\cite{cai2024toolmakers} & \cmark & \xmark & \xmark \\
      \ours (ours) & \cmark & \cmark & \cmark \\
      \bottomrule
    \end{tabular}
  }
  \caption{Comparison of tool creation methods. \emph{OS interaction} refers to the ability to interact with the \glsentrylong{os} (\eg read/write files, run commands, web browsing). \emph{Complex tasks} require installing and using external dependencies (\eg libraries, model weights).}
  \label{tab:tool-learning}
\end{table}

\paragraph{Benchmarks}
Various benchmarks have been proposed specifically for tool creation, and software engineering more generally.
Code generation benchmarks~\cite{zhuo2024bigcodebench,jain2024livecodebench} assess the ability of \glspl{llm} to generate Python functions for narrowly defined tasks (\eg simple mathematical problems) using the Python standard library.
Tool creation benchmarks extend this idea, enabling the \gls{llm} to decide the signature of the Python function in addition to generating the implementation itself~\cite{yuan2024craft,qian2023creator,cai2024toolmakers}.
Yet, these existing code generation and tool creation benchmarks are limited to simple Python functions -- they cannot install dependencies or directly interact with the \gls{os}.

On the other hand, software engineering benchmarks assess \gls{llm} agents for solving GitHub issues~\cite{jimenez2024swebench}, creating ML models~\cite{tang2024mlbench,chan2024mlebench} and performing repository-level scientific tasks~\cite{majumder2024discoverybench,chen2025scienceagentbench,bogin2024super,liu2024repobench}.
However, these benchmarks focus on performing \emph{particular tasks}, as opposed to creating a reusable tool to solve a class of problems.

We combine both streams (tool creation and software engineering) by proposing a benchmark focused on real-world multi-step scientific tasks that requires agents to (i) autonomously install necessary dependencies (as opposed to implementing simple Python functions), and (ii) produce a reusable tool that can be applied with different inputs (as opposed to solving a single task instance).

\begin{figure*}[t]
  \includegraphics[width=\linewidth]{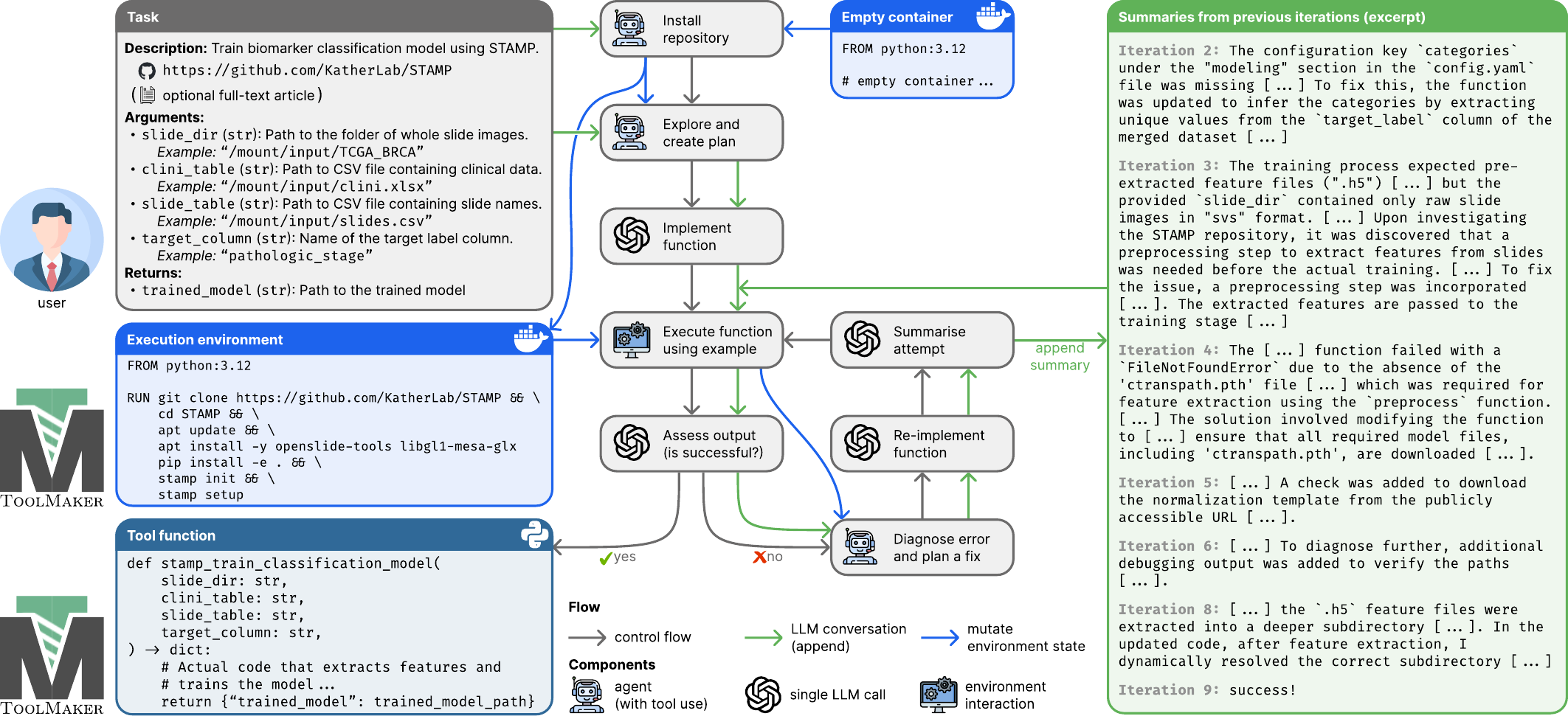}
  \caption{\textbf{\Ours workflow.}
    Given a task description, a scientific paper, and its associated code repository, \ours generates an executable tool that enables a downstream \gls{llm} agent to perform the described task.
  }
  \label{fig:overview}
\end{figure*}

\section{\Ours}
We design \ours to autonomously convert stand-alone code repositories from scientific publications into \gls{llm}-compatible tools. Each tool should complete a specific, user-defined task. To do so, we require a minimal \emph{tool definition} (see \cref{fig:tool_creation}, top), consisting of:
\begin{enumerate}[noitemsep,topsep=0pt,label=\arabic*)]
  \item a concise textual description of the task,
  \item GitHub URL of the associated repository, and
  \item a list of required input arguments, including an example value for each argument.
\end{enumerate}

This tool definition could in principle be represented as the signature of a Python function with a docstring, like in existing code generation tasks~\cite{zhuo2024bigcodebench,jain2024livecodebench}.
However, unlike previous work, we require the \gls{llm} to not only implement the function, but also to \emph{set up the environment} wherein the function will be executed.
The latter is necessary due to the complexity of our tasks which require \eg installing external dependencies, downloading models, and setting up configurations while considering system and hardware specifications.

We structure \ours as an \emph{agentic workflow} (\cref{fig:overview}) consisting of two stages: environment setup and tool implementation.
During environment setup, \ours produces a reproducible ``snapshot'' of the system (a Docker image) wherein the tool will run.
Then, \ours generates a Python function that implements the  desired task.

\def\conversation{\textcolor{llmflow}{h}}
\def\setofconversations{\textcolor{llmflow}{\mathcal{H}}}
\def\llmcall{\textcolor{llmflow}{\ell}}
\def\message{\textcolor{llmflow}{m}}
\def\setofmessages{\textcolor{llmflow}{\mathcal{M}}}
\def\setofreturns{\mathcal{R}}
\def\setofllmcalls{\textcolor{llmflow}{\mathcal{L}}}

\def\envstate{\textcolor{environmentflow}{e}}
\def\setofenvstates{\textcolor{environmentflow}{\mathcal{E}}}

\newcommand{\action}{\textcolor{environmentflow}{a}}
\def\setofactions{\textcolor{environmentflow}{\mathcal{A}}}
\def\agent{\textcolor{agentflow}{\ensuremath{\pi}}}
\def\observation{\textcolor{environmentflow}{o}}
\def\setofobservations{\textcolor{environmentflow}{\mathcal{O}}}
\def\workflowstate{s}
\def\setofworkflowstates{\mathcal{S}}

\subsection{Workflow components}

We define the \emph{state of the workflow} at any point in time to be a pair
\begin{equation*}
  s = \bigl(\conversation,\;\envstate\bigr) 
  \; \in \; \setofconversations \times \setofenvstates.
\end{equation*}
Here, $\conversation \in \setofconversations$ is the \emph{conversation history} (the ordered sequence of messages from the user, tools, and the \gls{llm}), and $\envstate \in \setofenvstates$ is the \emph{environment state} (represented by a checkpointed Docker container).

\Ours is built out of fundamental \emph{components}, each viewed as a function that acts on the workflow state as
\begin{equation*}
  \setofworkflowstates \;\mapsto\; \setofworkflowstates \times \setofreturns,
\end{equation*}
where $\setofworkflowstates = \setofconversations \times \setofenvstates$ is the space of all possible workflow states, and $\setofreturns \supseteq \setofmessages \cup \setofobservations$ is the set of possible returns (e.g.\ a newly generated message in $\setofmessages$ or an environment observation in $\setofobservations$). 
We distinguish three types of components:
\textbf{\icon{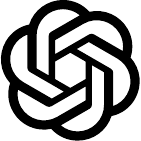}\:\gls{llm} calls} ($\setofconversations \mapsto \setofconversations \times \setofmessages$),
\textbf{\icon{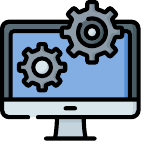}\:environment interactions} ($\setofenvstates \mapsto \setofenvstates \times \setofobservations$), and 
\textbf{\icon{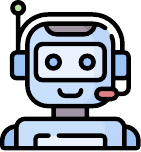}\:agents} ($\setofconversations \times \setofenvstates \mapsto \setofconversations \times \setofenvstates \times \setofreturns$).

\subsubsection{\icon{images/icons/llm_call.pdf}\:\gls{llm} calls} 
\Pgls{llm} can be viewed as a function
\begin{equation*}
  LLM : \setofconversations \;\to\; \setofmessages,
\end{equation*}
which, given a conversation history, produces a single new message. As a \ours workflow component, an \gls{llm} call $\llmcall : \setofconversations \to \setofconversations \times \setofmessages$ takes the workflow state's conversation history $\conversation$, appends $LLM(\conversation)$, and returns the new message:
\begin{equation*}
  \conversation
  \;\mapsto\;
  (\conversation \oplus LLM(\conversation),\; LLM(\conversation)).
\end{equation*}
\Glspl{llm} calls thus only update the conversation and do not modify the environment.
We use OpenAI's \texttt{gpt-4o-2024-08-06} model for the \gls{llm} calls.

\subsubsection{\icon{images/icons/environment_interaction.pdf}\:Environment interactions}
An environment interaction is any action $\action\in\setofactions$ that can read from or write to the environment state $\envstate$. We may thus model it by
\begin{equation*}
  \envstate
  \;\mapsto\;
  (\envstate',\;\observation),
\end{equation*}
where $\envstate'$ is the updated environment state, and $\observation\in\setofobservations$ is the observation produced by the action. 

The set of environment actions are
\begin{equation*}
  \setofactions = \adjustbox{max width=.8\linewidth}{$\displaystyle
  \left\{ 
    \begin{array}{l}
    \icon{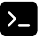}\:\textsc{run\_bash\_command},
    \icon{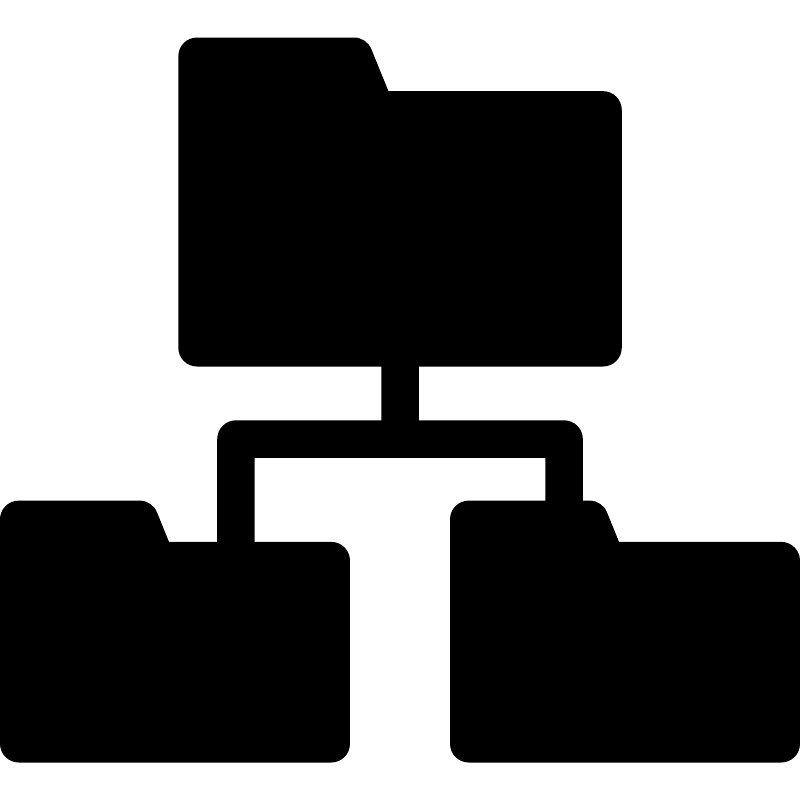}\:\textsc{list\_directory}, \\
    \icon{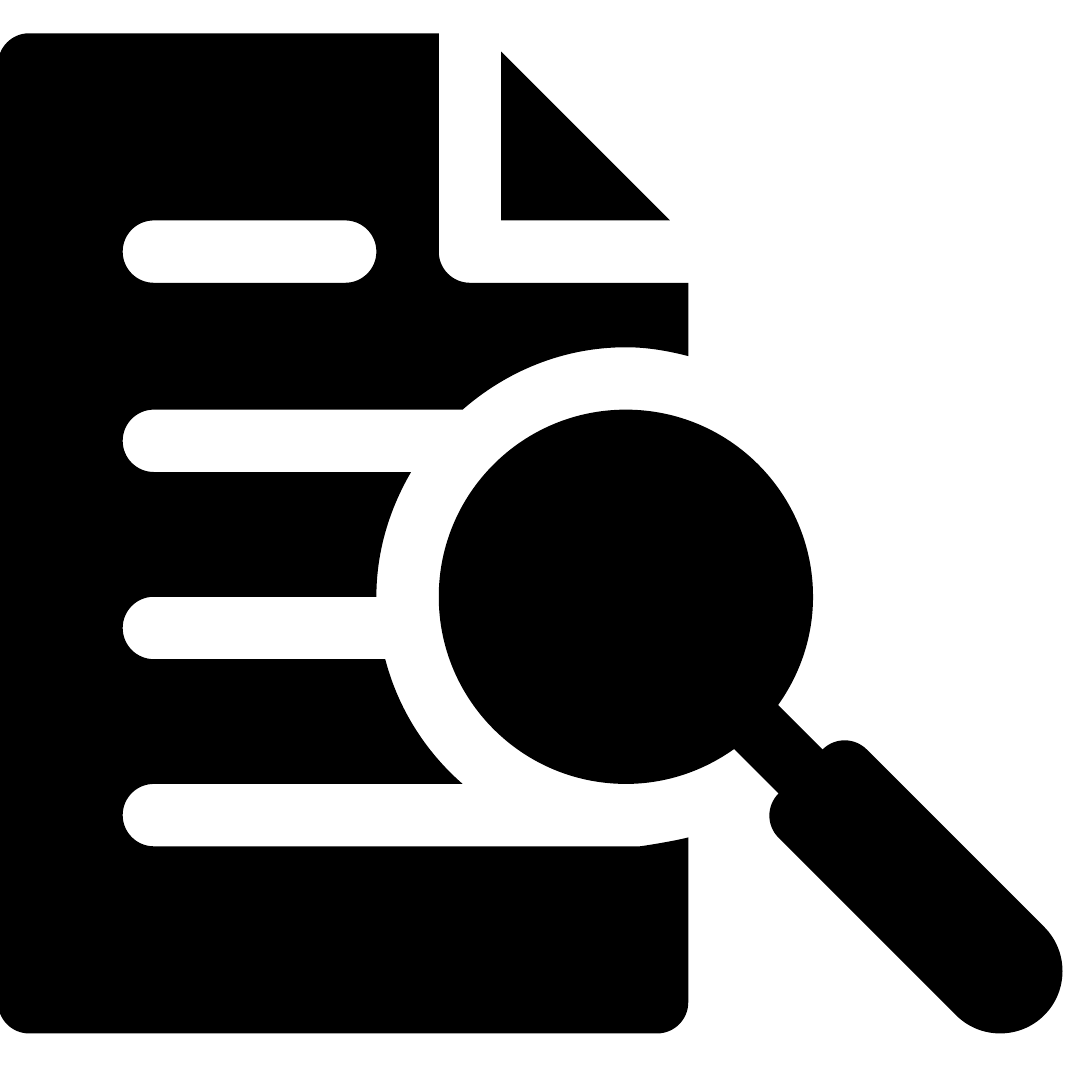}\:\textsc{read\_file},
    \icon{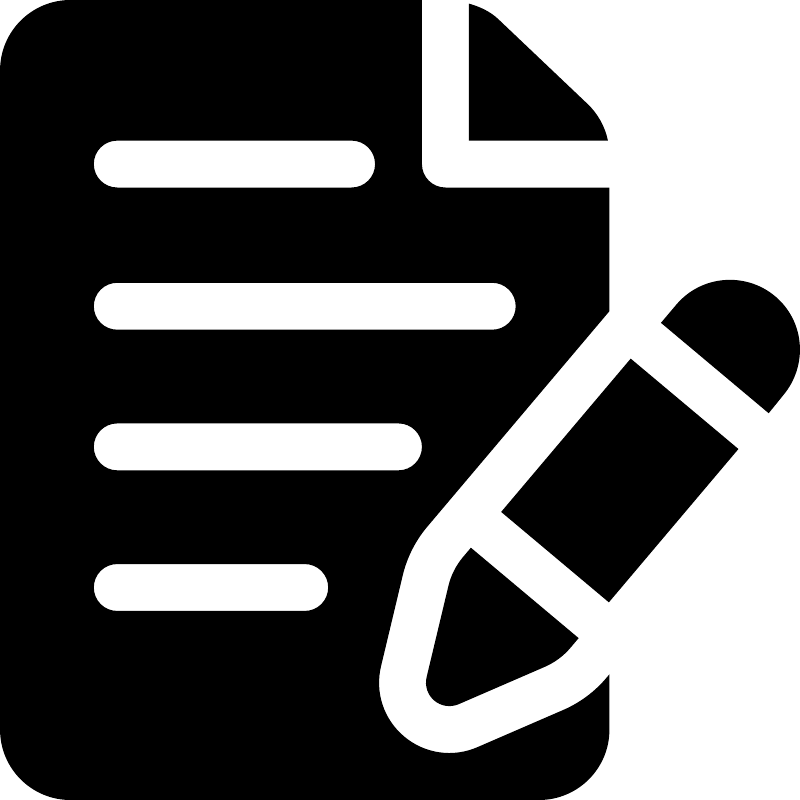}\:\textsc{write\_file},
    \icon{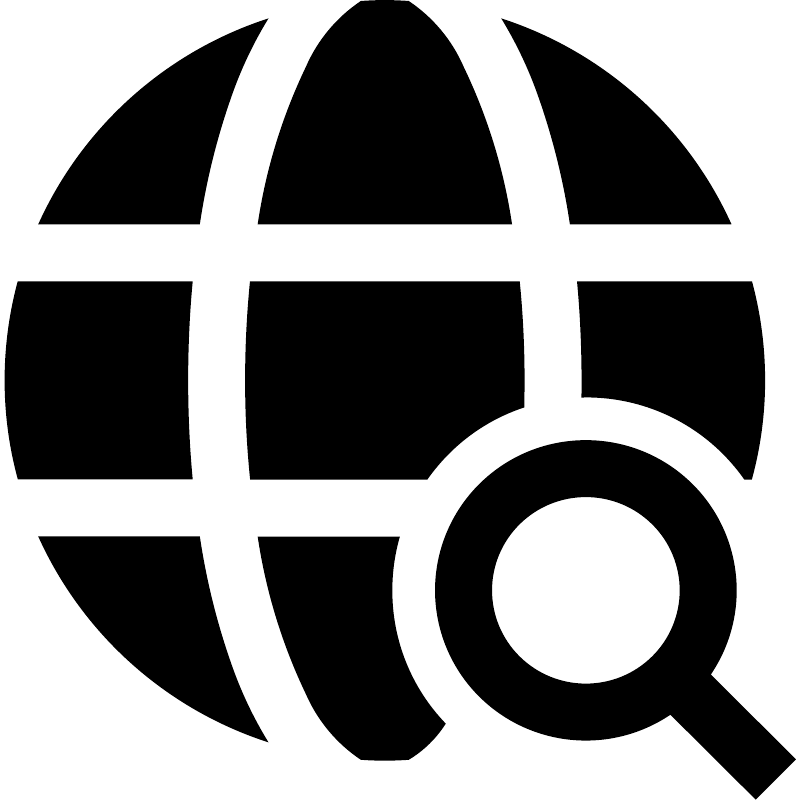}\:\textsc{browse}, \\
    \icon{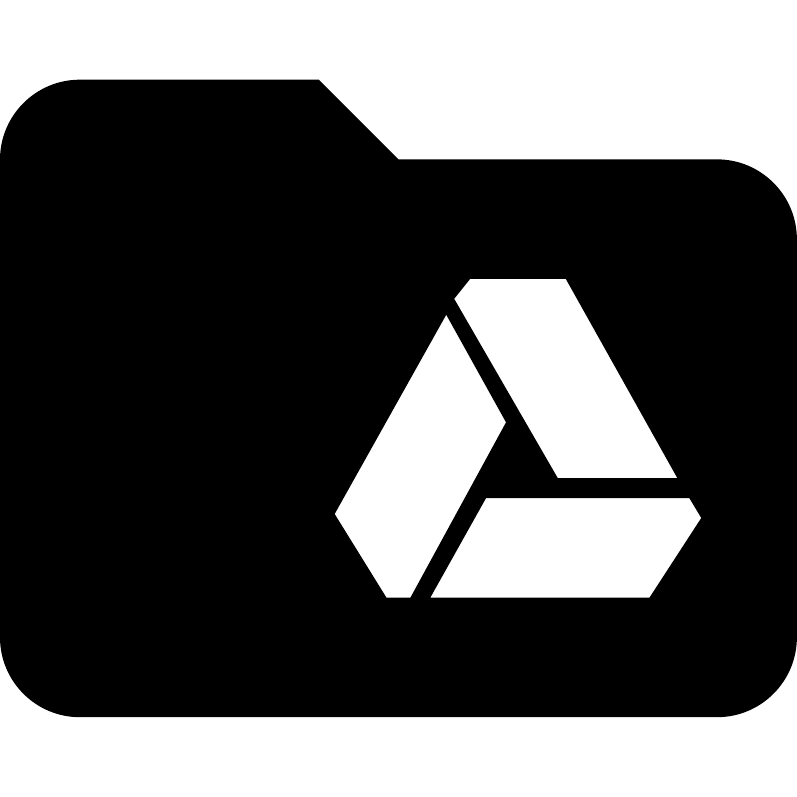}\:\textsc{google\_drive\_list\_folder}, \\
    \icon{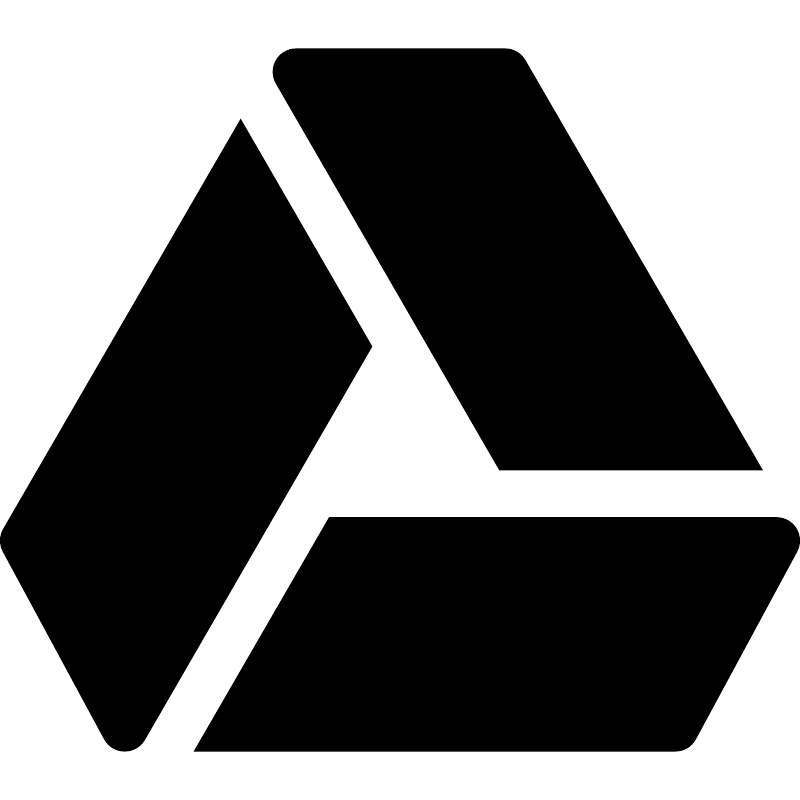}\:\textsc{google\_drive\_download\_file}, \\
    \icon{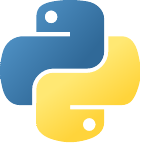}\:\textsc{run\_implementation}
    \end{array}
  \right\}
  $}.
\end{equation*}
We distinguish between \emph{read-only} actions and \emph{write} actions~\cite{huyen2024aiengineering}.
While read-only actions $\setofactions_{r} = \{\icon{images/icons/read_file.pdf}, \icon{images/icons/list_directory.pdf}, \icon{images/icons/browse.pdf}, \icon{images/icons/google_drive_folder.pdf}\}$ have $\envstate' = \envstate$, write actions $\setofactions_{w} = \{\icon{images/icons/terminal.pdf}, \icon{images/icons/write_file.pdf}, \icon{images/icons/google_drive.pdf}, \icon{images/icons/python.pdf}\}$ may modify $\envstate$.

The \icon{images/icons/python.pdf}\:\textsc{run\_implementation} action is a special action that allows \ours to execute a candidate tool implementation.

\subsubsection{\icon{images/icons/agent.pdf}\:Agents} 
\label{sec:agents}

\begin{figure}[h]
  \centering
  \includegraphics[width=.8\linewidth]{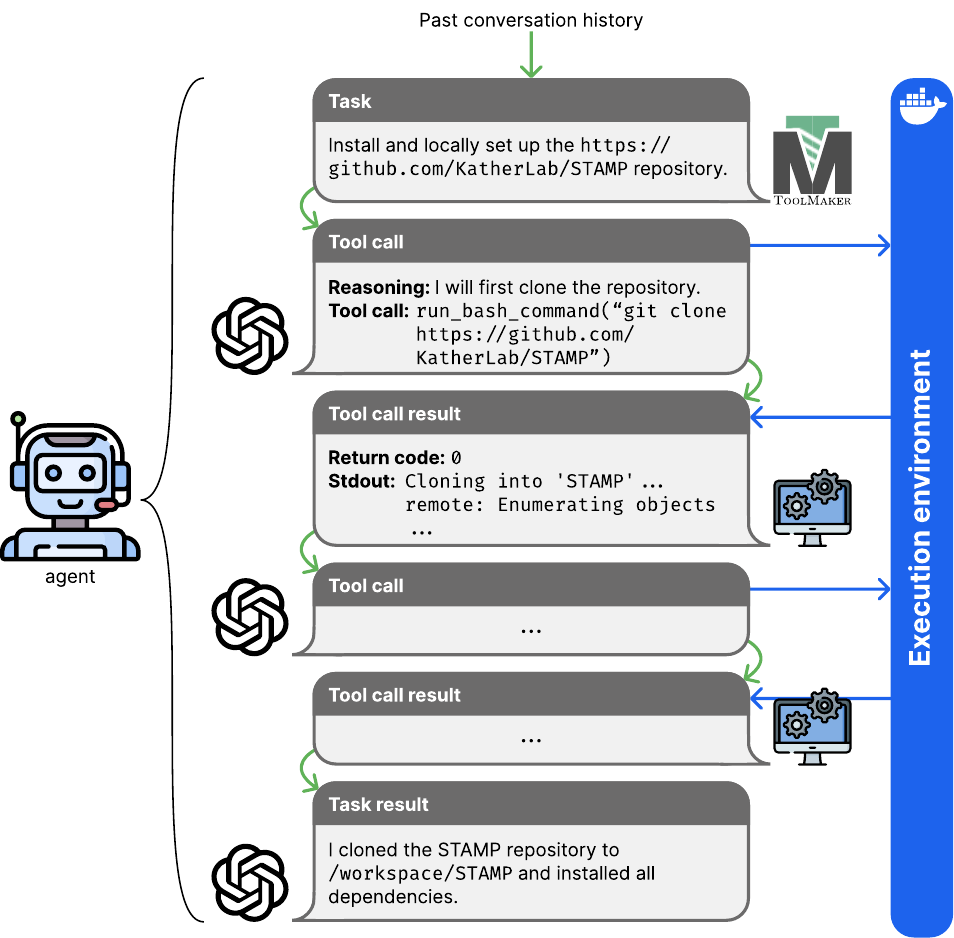}
  \caption{An agent uses a tool-augmented \gls{llm} to perform a specific sub-task, and returns the result. Messages are \textcolor{llmflow}{appended} to the conversation history, and tool calls enable the agent to \textcolor{environmentflow}{interact} with the environment.}
  \label{fig:sub_agent}
\end{figure}

An \emph{agent} $\agent$, illustrated in \cref{fig:sub_agent}, chains multiple \gls{llm} calls and environment interactions to accomplish a specific sub-task which is specified by a high-level instruction, $\message_{\agent} \in \setofmessages$, \eg ``install this repository and its dependencies''.

Formally, an agent $\agent$ maps the current workflow state $s=(\conversation, \envstate)$ to a new state $s_T=(\conversation_T, \envstate_T)$ and return value $r\in\setofreturns$:
\begin{equation*}
  (\conversation, \envstate) \;\mapsto\; (\conversation_T, \envstate_T, r).
\end{equation*}

The agent follows a sequence of state transitions
\begin{equation*}
  s_0 \;\to\; s_1 \;\to\; \cdots \;\to\; s_T,
\end{equation*}
where each state $s_t = (\conversation_t, \envstate_t) \in \setofworkflowstates$.
At step $t=0$, the agent receives the \emph{initial} state
\begin{equation*}
  s_0 \;=\;
  \bigl(\conversation \oplus \message_{\agent},\;\envstate\bigr).
\end{equation*}

At each step $t$, the agent employs a special \emph{tool-augmented} \gls{llm}, denoted
\begin{equation*}
  LLM_{\agent} : \setofconversations
  \;\to\;
  \setofactions_{\agent} \;\cup\; \setofreturns,
\end{equation*}
which, given the current conversation $\conversation_t$, either outputs an \textbf{action} $\action_t \in \setofactions_{\agent}$ (a tool call) or the \textbf{final result} $r \in \setofreturns$ of the sub-task. Here, $\setofactions_{\agent} \subseteq \setofactions \setminus \{\icon{images/icons/python.pdf}\:\textsc{run\_implementation}\}$ excludes directly running candidate tool implementations, as this is a separate step in the \ours workflow.
We implement the choice between $\setofactions_{\agent}$ and $\setofreturns$ using OpenAI's function calling and structured output \glspl{api} respectively~\cite{openai2025docs}.

If the \gls{llm} proposes an action $\action_t=LLM_{\agent}(\conversation_t) \in \setofactions$, we execute $\action_t$ on the current environment to obtain the observation and updated environment state $(\envstate_{t+1}, \observation_t)=\action_t(\envstate_t)$. We then append both the tool call and its observation to the conversation, forming the new state
\begin{equation*}
  s_{t+1} \;=\;
  (\conversation_t \oplus \action_t \oplus \observation_t,\; \envstate_{t+1}).
\end{equation*}
If instead $LLM_{\agent}(\conversation_t)$ outputs a final result $r \in \setofreturns$, the agent terminates and returns $s_T=(\conversation_t, \envstate_t, r)$.

\subsection{\Ours workflow}
\label{sec:toolmaker_workflow}
In this section, we describe our workflow in detail, which at a high level is illustrated in \cref{fig:overview}, and in pseudocode in \cref{alg:toolmaker}, using the three types of components (\icon{images/icons/llm_call.pdf}\:\gls{llm} calls, \icon{images/icons/environment_interaction.pdf}\:environment interactions, and \icon{images/icons/agent.pdf}\:agents) introduced above.

\begin{algorithm}[t]
  \caption{\Ours workflow.}
  \label{alg:toolmaker}
  \small
  \begin{algorithmic}[1]
    \Require Tool definition $\message_{\text{tool}}$, initial environment $\envstate_\emptyset \in \setofenvstates$
    \State $\conversation_\emptyset \gets \{\message_{\text{tool}}\}$ \Comment{initialise conversation history}
    \State $\conversation, \envstate, r \gets \textsc{\icon{images/icons/agent.pdf}\:install\_repository}(\conversation_\emptyset, \envstate_\emptyset)$ \label{line:install_repository}
    \State $\bar{\envstate} \gets \envstate$ \Comment{snapshot of installed environment state} \label{line:snapshot_envstate}
    \State $\conversation, \envstate, r \gets \textsc{\icon{images/icons/agent.pdf}\:explore}(\conversation_\emptyset, \bar{\envstate})$ \label{line:explore}
    \State $\conversation, \message \gets \textsc{\icon{images/icons/llm_call.pdf}\:plan}(\conversation)$ \label{line:plan}
    \State $\bar{\conversation} \gets \conversation$ \label{line:snapshot_conversation_history}\Comment{snapshot of conversation history}
    \State $\conversation, \message_\text{code} \gets \textsc{\icon{images/icons/llm_call.pdf}\:implement}(\conversation)$ \label{line:implement}
    \State $\sigma \gets \emptyset$
    \While{true} \label{line:loop}
      \State $\envstate \gets \bar{\envstate}$ \Comment{restore installed environment state}\label{line:restore_envstate}
      \State $\conversation \gets \bar{\conversation} \oplus \sigma \oplus \message_\text{code}$ \label{line:restore_conversation_history} \Comment{restore conversation history}
      \State $\envstate, \observation \gets \textsc{\icon{images/icons/environment_interaction.pdf}\:run\_implementation}(\envstate, \message_\text{code})$ \label{line:run_implementation}
      \State $\conversation, \message \gets \textsc{\icon{images/icons/llm_call.pdf}\:assess\_tool\_output}(\conversation \oplus \observation)$ \label{line:assess_tool_output}
      \If{$\message$ is successful}
        \State \Return $\bar{\envstate}, \message_\text{code}$
      \EndIf
      \State $\conversation, \envstate, r \gets \textsc{\icon{images/icons/agent.pdf}\:diagnose\_error}(\conversation \oplus \observation, \envstate)$ \label{line:diagnose_error}
      \State $\conversation, \message_\text{code} \gets \textsc{\icon{images/icons/llm_call.pdf}\:reimplement}(\conversation)$ \label{line:reimplement}
      \State $\conversation, \message_\text{summary} \gets \textsc{\icon{images/icons/llm_call.pdf}\:summarise}(\conversation)$ \label{line:summarise}
      \State $\sigma \gets \sigma \oplus \message_\text{summary}$ \label{line:update_sigma}
    \EndWhile
  \end{algorithmic}
\end{algorithm}

\Ours's initial conversation history $\conversation_\emptyset$ is a system prompt that contains the tool definition $\message_{\text{tool}}$. 
We provide the full prompts in \cref{app:prompts}.

\paragraph{Environment setup}
To obtain the state of the execution environment necessary for the tool to execute, we employ the \icon{images/icons/agent.pdf}\:\textsc{install\_repository} agent (line~\ref{line:install_repository}) that is instructed to install and set up the repository.
This agent clones and explores the repository, reads documentation, and downloads any dependencies it deems necessary such as models, datasets, and libraries. Each of these steps involve planning and learning from previous observations such as error logs arising during execution.

The agent begins with a clean environment state $\envstate_\emptyset$ (a \mbox{\texttt{python:3.12}} Docker image).
Importantly, we record all write actions ($\setofactions_{w}$) that the agent performs.
Since each of these actions may be expressed as a bash command, we simply concatenate their bash representations to obtain the environment definition in the form of a bash script or Dockerfile.

\paragraph{Initial implementation}
We first instruct an agent (\icon{images/icons/agent.pdf}\:\textsc{explore}) to explore the repository and gather all information necessary to implement the tool.
Note that we do not carry over the conversation history from the previous stage, in order to not pollute the context with a large number of messages (by calling \icon{images/icons/agent.pdf}\:\textsc{explore} on $\conversation_\emptyset$, not $\conversation$ on line~\ref{line:explore}).

Next we perform an \gls{llm} call (\icon{images/icons/llm_call.pdf}\:\textsc{plan}) to create a step-by-step plan for the implementation.
We keep all messages (including actions and observations) in the conversation history, so this information can be used to create the plan.

Then, we instruct the \gls{llm} (\icon{images/icons/llm_call.pdf}\:\textsc{implement}) to write the Python code for the tool based on the plan, producing our first \emph{candidate implementation}.

\paragraph{Closed-loop self-improvement}
Now, we enter the closed-loop self-improvement phase.
First, we reset the execution environment to the \emph{environment definition} $\bar{\envstate}$ because the agent may have performed write actions in the past.
We also restore the conversation history to immediately after generating the implementation plan, but include summaries of past appempts (described later).

After running the candidate Python function in the execution environment using the example invocation provided in the tool definition (line~\ref{line:run_implementation}), we instruct the \gls{llm} to assess whether the execution was successful (\icon{images/icons/llm_call.pdf}\:\textsc{assess\_tool\_output}).
Specifically, we ask the \gls{llm} to check whether the result returned by the tool is in line with the task description (\ie if the result is plausible), and whether the standard output and standard error streams contain any indications of errors.
If the \gls{llm} deemed tool execution successful, we have arrived at our final tool implementation, and exit the loop.
Otherwise, we continue the self-improvement loop.

Next, we instruct the \icon{images/icons/agent.pdf}\:\textsc{diagnose\_error} agent to gather information about the error in order to diagnose its root cause and formulate a plan to fix it.
Importantly, we do not reset the execution environment -- the agent is able to check intermediate files and outputs created during tool execution.

Then, the \gls{llm} re-implements the tool based on the current implementation, error diagnosis, and plan to fix the error (\icon{images/icons/llm_call.pdf}\:\textsc{reimplement}). 
Finally, we ask the \gls{llm} to summarise the current step (\icon{images/icons/llm_call.pdf}\:\textsc{summarise}), and append this summary to the conversation history for the next iteration.

\subsection{Execution environment\:\icon{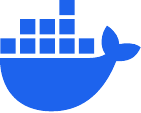}}
An important implementation detail is the \emph{execution environment}, which is the environment in which (i) actions ($\setofactions$) are performed throughout the \ours workflow, and (ii) wherein the final tool created by \ours will be executed.

The execution environment itself is \emph{stateful}. Specifically, write actions $\setofactions_{w} = \{\icon{images/icons/terminal.pdf}, \icon{images/icons/write_file.pdf}, \icon{images/icons/google_drive.pdf}, \icon{images/icons/python.pdf}\}$ may mutate environment state.
However, we require the ability to roll back to previous states, \eg
on line~\ref{line:restore_envstate} of \cref{alg:toolmaker}, the execution environment is restored to the ``freshly installed'' state $\bar{\envstate}$.
Furthermore, the execution environment should be sandboxed from the host system (for security reasons), and it should be reproducible (so the generated tool can be executed on any machine).

We satisfy these requirements by implementing the execution environment as a Docker container that \ours controls via \pgls{http} server running inside the container, which can run the pre-defined actions $\setofactions$.
State restoration is achieved via Docker's checkpointing functionality.

\section{Benchmark}

\begin{table*}
  \centering
  \adjustbox{max width=\linewidth}{
    \begin{tabular}{ll|rrrrr|rrrrr}
\toprule
&  & \multicolumn{5}{c}{\textbf{\ours (ours)}} & \multicolumn{5}{c}{\textbf{OpenHands~\cite{wang2024openhands}}} \\
\multicolumn{2}{c|}{\textbf{Task}} & \textbf{Invoc.} & \textbf{Tests} & \textbf{Cost} & \textbf{Actions} & \textbf{Tokens} & \textbf{Invoc.} & \textbf{Tests} & \textbf{Cost} & \textbf{Actions} & \textbf{Tokens} \\
\midrule
\multirow[t]{6}{*}{Pathology} & \texttt{conch\_extract\_features}~\cite{lu2024conch} & \cellcolor{cellgreen} 3/3 & \cellcolor{cellgreen} 9/9 & \$0.35 & 15 ($1_\circlearrowleft$) & 171,226 & \cellcolor{cellgreen} 3/3 & \cellcolor{cellgreen} 9/9 & \$0.08 & 5 & 51,701 \\
 & \texttt{musk\_extract\_features}~\cite{xiang2025musk} & \cellcolor{cellgreen} 3/3 & \cellcolor{cellgreen} 6/6 & \$1.19 & 29 ($6_\circlearrowleft$) & 696,386 & \cellcolor{cellred} \errorinstallfailed & \cellcolor{cellred} \errorinstallfailed & \$0.15 & 7 & 97,386 \\
 & \texttt{pathfinder\_verify\_biomarker}~\cite{liang2023pathfinder} & \cellcolor{cellred} 0/2 & \cellcolor{cellyellow} 4/6 & \$0.61 & 27 ($1_\circlearrowleft$) & 356,825 & \cellcolor{cellred} 0/2 & \cellcolor{cellyellow} 4/6 & \$0.08 & 6 & 49,414 \\
 & \texttt{stamp\_extract\_features}~\cite{elnahhas2024stamp} & \cellcolor{cellgreen} 3/3 & \cellcolor{cellgreen} 12/12 & \$1.12 & 20 ($4_\circlearrowleft$) & 631,138 & \cellcolor{cellred} 0/3 & \cellcolor{cellyellow} 3/12 & \$0.07 & 6 & 42,793 \\
 & \texttt{stamp\_train\_classification\_model}~\cite{elnahhas2024stamp} & \cellcolor{cellgreen} 3/3 & \cellcolor{cellgreen} 9/9 & \$2.27 & 33 ($9_\circlearrowleft$) & 1,249,521 & \cellcolor{cellred} 0/3 & \cellcolor{cellred} 0/9 & \$0.15 & 8 & 87,915 \\
 & \texttt{uni\_extract\_features}~\cite{chen2024uni} & \cellcolor{cellgreen} 3/3 & \cellcolor{cellgreen} 9/9 & \$0.61 & 16 ($4_\circlearrowleft$) & 326,806 & \cellcolor{cellred} \errorinstallfailed & \cellcolor{cellred} \errorinstallfailed & \$0.25 & 10 & 177,119 \\
\hline
\multirow[t]{2}{*}{Radiology} & \texttt{medsam\_inference}~\cite{ma2024medsam} & \cellcolor{cellgreen} 3/3 & \cellcolor{cellgreen} 6/6 & \$0.96 & 18 ($6_\circlearrowleft$) & 508,954 & \cellcolor{cellred} \errorinstallfailed & \cellcolor{cellred} \errorinstallfailed & \$0.07 & 5 & 41,096 \\
 & \texttt{nnunet\_train\_model}~\cite{isensee2020nnunet} & \cellcolor{cellred} 0/2 & \cellcolor{cellred} 0/4 & \$2.90 & 35 ($9_\circlearrowleft$) & 1,792,291 & \cellcolor{cellred} 0/2 & \cellcolor{cellred} 0/4 & \$0.12 & 8 & 79,231 \\
\hline
\multirow[t]{2}{*}{Omics} & \texttt{cytopus\_db}~\cite{kunes2023cytopus} & \cellcolor{cellgreen} 3/3 & \cellcolor{cellgreen} 12/12 & \$0.41 & 10 ($3_\circlearrowleft$) & 185,912 & \cellcolor{cellred} \errorinstallfailed & \cellcolor{cellred} \errorinstallfailed & \$0.36 & 8 & 236,217 \\
 & \texttt{esm\_fold\_predict}~\cite{verkuil2022esm1,hie2022esm2} & \cellcolor{cellyellow} 2/3 & \cellcolor{cellyellow} 13/15 & \$0.66 & 20 ($1_\circlearrowleft$) & 336,754 & \cellcolor{cellred} \errorinstallfailed & \cellcolor{cellred} \errorinstallfailed & \$0.11 & 6 & 69,493 \\
\hline
\multirow[t]{5}{*}{Other} & \texttt{flowmap\_overfit\_scene}~\cite{smith2024flowmap} & \cellcolor{cellgreen} 2/2 & \cellcolor{cellgreen} 6/6 & \$0.70 & 18 ($5_\circlearrowleft$) & 358,552 & \cellcolor{cellred} \errorinstallfailed & \cellcolor{cellred} \errorinstallfailed & \$0.36 & 15 & 250,787 \\
 & \texttt{medsss\_generate}~\cite{jiang2025medsss} & \cellcolor{cellgreen} 3/3 & \cellcolor{cellgreen} 6/6 & \$0.53 & 25 ($3_\circlearrowleft$) & 282,771 & \cellcolor{cellgreen} 3/3 & \cellcolor{cellgreen} 6/6 & \$0.15 & 10 & 104,505 \\
 & \texttt{modernbert\_predict\_masked}~\cite{warner2024modernbert} & \cellcolor{cellgreen} 3/3 & \cellcolor{cellgreen} 9/9 & \$0.66 & 20 ($4_\circlearrowleft$) & 356,228 & \cellcolor{cellred} \errorinstallfailed & \cellcolor{cellred} \errorinstallfailed & \$0.13 & 10 & 82,930 \\
 & \texttt{retfound\_feature\_vector}~\cite{zhou2023retfound} & \cellcolor{cellgreen} 3/3 & \cellcolor{cellgreen} 6/6 & \$0.97 & 31 ($5_\circlearrowleft$) & 561,936 & \cellcolor{cellred} 0/3 & \cellcolor{cellred} 0/6 & \$0.08 & 4 & 46,521 \\
 & \texttt{tabpfn\_predict}~\cite{hollmann2025tabpfn} & \cellcolor{cellgreen} 3/3 & \cellcolor{cellgreen} 9/9 & \$0.23 & 10 ($1_\circlearrowleft$) & 95,257 & \cellcolor{cellgreen} 3/3 & \cellcolor{cellgreen} 9/9 & \$0.07 & 4 & 36,320 \\
\hline
\bottomrule
\end{tabular}

  }
  \caption{Performance of the tools created by \ours and the OpenHands baseline~\cite{wang2024openhands} on the benchmark tasks.
    \errorinstallfailed indicates that the environment installation failed.
    We use $\circlearrowleft$ to indicate the number of self-correcting iterations. \coloredbox{cellgreen}{Green} cells indicate that the tool implementation is correct (all unit tests pass), \coloredbox{cellyellow}{yellow} indicates that at least one unit test failed, and \coloredbox{cellred}{red} indicates that all unit tests failed.}
  \label{tab:results}
\end{table*}

To evaluate our approach, we collect a dataset of 15 diverse tasks spanning multiple scientific disciplines, which we refer to as \ourbenchmark. 
The tasks were curated in close collaboration with researchers in medicine and life sciences to reflect realistic problems in these fields, with a focus on the medical domain (pathology, radiology, omics), while also including some tasks from other areas such as 3D vision, imaging, tabular data analysis, and natural language processing to ensure broader coverage of real-world scientific challenges.

Before including a task in \ourbenchmark, we manually implemented the intended tool using the associated GitHub repository to ensure the task is well-defined and solvable.
This vetting process gave us confidence that each task is meaningful, correctly specified, and feasible.
The resulting benchmark covers a range of difficulty levels, from simple tasks that can be achieved by calling an existing method, to more complex, multi-step tasks that require orchestrating multiple function calls, transforming data, and utilising \glspl{gpu}.

\paragraph{Task definitions}

As shown in \cref{fig:tool_creation} (top), each task definition consists of:
\begin{enumerate*}[label=(\roman*)]
  \item a one-sentence task description,
  \item the URL to the code repository,
  \item a list of input arguments, alongside an example invocation (see below), and
  \item a description of the expected output.
\end{enumerate*}
An overview of the tasks and associated papers can be found in \cref{tab:results}, and we provide a full list of all task definitions with their example invocations in \cref{app:benchmark}.

\paragraph{Invocations}
A task \emph{invocation} specifies a concrete value for each input argument, as well as \emph{external files and directories} that should be made accessible from within the execution environment during the invocation. Indeed, most tasks in \ourbenchmark require external files, \eg \texttt{stamp\_train\_classification\_model} takes an input dataset of \glspl{wsi} and a clinical data table, on which to train a classification model using the STAMP~\cite{elnahhas2024stamp} pipeline. Analysing and utilising datasets is a fundamental aspect of many real-world scientific tasks, which is why \ourbenchmark explicitly supports this functionality, unlike many existing code generation benchmarks~\cite{zhuo2024bigcodebench,jain2024livecodebench}.

Each task definition includes a single example invocation, which may be used in the tool creation process. Crucially, this specification does not include the expected return value, as the goal is to autonomously implement and execute the task without prior knowledge of the correct output.

\paragraph{Assessing correctness}
\Ourbenchmark specifies 2-3 additional test invocations per task, which are different to the example invocation and held-out from the tool creation process.
We purposefully chose different input argument values for the test invocations (different datasets, images, paths, \etc) to assess whether the implementations would generalise to other inputs, and to ensure that implementations did not `hard-code' the example invocation.

For each test invocation, \ourbenchmark includes unit tests to assess whether the tool produces the expected output by checking various properties of the return value and output files.
We opted for unit tests over simple equality checks (\eg strict or near-exact matches to reference outputs, as used in previous benchmarks~\cite{bogin2024super}) because unit tests can accommodate more complex criteria, such as verifying the shape of generated feature vectors or checking that a segmentation model produces plausibly sized masks.
Specifically, we employ unit tests to verify correctness through assertions on: \emph{structure} (dimensions and types of return values), \emph{values} (range, accuracy, and statistical properties of return values), \emph{files} (existence, format, and content of files produced by the tool, if applicable), and \emph{execution} (errors/crashes).

To ensure an unbiased assessment of tool implementations, the unit tests and test invocations are used strictly for evaluation and are not available during tool creation. 
\ourbenchmark comprises 15 tasks, with a total of 42 test invocations (average 2.8 per task) and 124 unit tests (average 8.3 per task). 
We consider a tool implementation correct only if it passes all unit tests of its test invocations.

\section{Results}

\Ourbenchmark can evaluate any ``tool maker'' that produces an environment definition~\icon{images/icons/docker.pdf} and a tool implementation~\icon{images/icons/python.pdf}.
However, to the best of our knowledge, no existing approaches are specifically designed to address the ``paper repository $\to$ \gls{llm} tool'' problem.
In order to nonetheless facilitate comparison with prior work, we adapt the OpenHands~\cite{wang2024openhands} to this setting. 
OpenHands is a software engineering agent that achieves SOTA performance on SWE-bench~\cite{jimenez2024swebench}.
We instruct OpenHands to generate the same artifacts as \ours: an environment definition~\icon{images/icons/docker.pdf} (expressed as a bash script to be run in a fresh \texttt{python:3.12} Docker image to create the environment state required for the tool to execute) and a tool implementation~\icon{images/icons/python.pdf} (a Python function). 
To ensure a fair comparison, we reuse large parts of the \ours prompts in the prompts we supply to the OpenHands, and add additional instructions to encourage OpenHands to test the artifacts it creates.
We use \texttt{gpt-4o} for the OpenHands baseline, but also ablate the choice of \gls{llm} in \cref{sec:ablations}.
The full prompts for \ours and OpenHands are listed in \cref{app:prompts,app:openhands_prompts}.

\paragraph{Performance}
In \cref{tab:results}, we report the performance of \ours and OpenHands on all tasks in \ourbenchmark, reporting correctness, cost, number of tokens, number of actions performed in the tool creation process (both stages), and the number of self-correcting iterations.
We consider a test invocation successful (``Tests'' column marked \coloredbox{cellgreen}{green}) if all of the unit tests that are associated with it pass.
Similarly, a tool implementation is correct (``Invoc.'' column marked \coloredbox{cellgreen}{green}) if \emph{all} of its test invocations are successful, \ie all of the unit tests associated with its test invocations pass.

\Ours significantly outperforms OpenHands, achieving an accuracy of 80\% (correctly implementing 12/15 tasks) while OpenHands was only able to correctly implement 20\% (3/15 tasks).

For the \texttt{esm\_fold\_predict}~\cite{verkuil2022esm1} task, \ours generates a partially correct implementation (\coloredbox{cellyellow}{yellow}) that passes two out of three test invocations.
The goal of this task is to predict the contact map of a protein from its sequence.
Upon inspection, we determined that the failed test invocation was different from the other invocations: it contained a mask token in the input sequence which was not present in the task definition's example invocation. 
However, when including such a mask token in the example invocation and re-running \ours, the tool implementation passed all test invocations.
This highlights that the example invocation in the task definition needs to be representative of the task. 

By contrast, OpenHands fails to generate correct tool implementations for most tasks, primarily due to errors at the environment setup stage. Nearly half of its environment definitions were invalid, causing installation scripts to crash before execution. Even among the tasks where OpenHands successfully generated an environment definition, only three implementations passed all unit tests.

This poor performance can largely be attributed to issues during environment setup.
Specifically, OpenHands often created installation scripts without testing them, omitted essential setup commands previously executed manually, or overlooked dependencies necessary for tool execution.
In contrast, \ours inherently avoids such pitfalls by automatically capturing every installation command and resetting the execution environment between iterations, ensuring reproducible and robust tool implementations.

\paragraph{Multi-step tools}
A remarkable feature of \ours is that it is able to create tools that require multiple steps to complete.
For example, the \texttt{stamp\_train\_classification\_model} task provides a dataset of pathology \glspl{wsi} and a table of clinical data, and requires the tool implementation to use the STAMP pipeline~\cite{elnahhas2024stamp} to train a classification model that predicts a specific biomarker from the \gls{wsi} images.
This task requires multiple steps to complete: 
after downloading and installing the STAMP repository and its dependencies, the tool implementation needs to use STAMP to
\begin{enumerate*}[label=(\arabic*)]
  \item perform feature extraction on the \gls{wsi} images, and
  \item train a classification model using the extracted features and the clinical data.
\end{enumerate*}
The self-correcting loop allows \ours to realise that it needs to perform feature extraction before it can train a classification model, and to subsequently implement the tool function to perform both steps, illustrated in \cref{fig:overview} (right).
For this particular task, \ours performs 9 self-correcting iterations, executing 33 actions in total, before arriving at the final implementation.

\paragraph{Cost}
\Ours performs an average of 21.8 actions during tool creation, costing on average \$0.94 per tool, while OpenHands performs 7.5 actions on average (\$0.15 per tool).
The three tools that OpenHands correctly implemented were among the cheapest for \ours, requiring the fewest actions and self-correcting iterations.
This shows OpenHands can implement very ``easy'' tools, but fails to generalise to more complex tasks.

\subsection{Ablations}
\label{sec:ablations}
\paragraph{Paper summaries}
Since each task is based on one or more research papers, we perform an ablation study to determine whether we can inject useful information from the papers into the tool creation process.
Instead of directly including the full paper text in the prompts which would require too many tokens, we first provide the full text to \texttt{gpt-4o} and instruct it to summarise it with respect to the task at hand.
Then, we provide these task-specific and paper-specific summaries in the prompts for \ours and OpenHands.

\begin{table}[t]
  \centering
  \adjustbox{max width=\linewidth}{
    \begin{tabular}{lrrrrr}
\toprule
\textbf{Method} & \textbf{Tools} & \textbf{Invoc.} & \textbf{Tests} & \textbf{Cost} & \textbf{Actions} \\
\midrule
{\ours}\textsuperscript{\textasteriskcentered} (ours) & 12/15 & 37/42 & 116/124 & \$0.94 & 21.8 \\
\qquad (with paper summary) & 11/15 & 34/42 & 113/124 & \$0.71 & 18.1 \\
\qquad (using o3-mini) & 9/15 & 28/42 & 107/124 & \$0.55 & 14.1 \\
\hline {OpenHands}\textsuperscript{\textasteriskcentered}~\cite{wang2024openhands} & 3/15 & 9/42 & 31/124 & \$0.15 & 7.5 \\
\qquad (with paper summary) & 3/15 & 9/42 & 31/124 & \$0.12 & 6.6 \\
\qquad (using o3-mini) & 1/15 & 2/42 & 15/124 & \$0.04 & 1.9 \\
\qquad (using Claude 3.5 Sonnet) & 2/15 & 6/42 & 19/124 & \$0.13 & 5.2 \\
\bottomrule
\end{tabular}

  }
  \caption{Ablation results. Rows marked with asterisk correspond to the results in \cref{tab:results}. We report the number of correct tools, invocations, and tests, as well as the per-tool average cost and number of actions performed.}
  \label{tab:ablations}
\end{table}

The results in \cref{tab:ablations} indicate that including paper summaries does not increase the performance of either approach.
However, it does decrease the average number of actions and, for \ours, the average number of self-correcting iterations required to create the tools.
For example, while \ours required 9 iterations (33 actions) to create the \texttt{stamp\_train\_classification\_model} tool, this decreased to only 5 iterations (15 actions) when using the paper summary (see \cref{app:extended_ablation_results}).

\paragraph{Choice of \gls{llm}}
We also evaluate \ours and OpenHands using OpenAI's \texttt{o3-mini} model instead of \texttt{gpt-4o}, and find that while this reduces cost, it also degrades performance in both cases.
Finally, since OpenHands achieved SOTA performance on SWE-bench~\cite{jimenez2024swebench} using Claude 3.5 Sonnet~\cite{anthropic_claude}, we re-run the OpenHands baseline using this model, but find that it performs worse than using \texttt{gpt-4o} (see \cref{tab:ablations}).

\section{Conclusion}

In this work, we showed that autonomous tool creation can go beyond simple Python functions and produce tools for real-world scientific tasks. We introduced \ours, a framework that autonomously transforms scientific code repositories into LLM-compatible tools, potentially drastically reducing the technical overhead in future for developing agents with specialised toolsets. 
In evaluations across multiple scientific domains, \ours surpassed the state-of-the-art software engineering agent, OpenHands, achieving 80\% accuracy. Additionally, we release \ourbenchmark as a comprehensive benchmark to spur further advancements in agentic tool creation.

We acknowledge that automated tool creation in life sciences carries significant risks that require careful consideration. The ability to autonomously implement complex biochemical tools could potentially be misused for creating harmful agents or bioweapons. Additionally, fully automated research systems might inadvertently generate dangerous compounds or protocols without proper oversight. These risks underscore the importance of developing robust safety measures and ethical guidelines alongside technical capabilities.
Nonetheless, by removing technical barriers to tool creation, \ours brings us closer to a future where the pace of scientific discovery is limited by computational capacity, not human resources.

\section*{Limitations}
While \ours addresses the challenge of tool creation, we acknowledge that fully autonomous scientific discovery remains constrained by physical experimentation. \Ours does not address this aspect, but we believe that with an increasing proportion of life science research being conducted \emph{in silico}, it provides a building block for autonomous scientific workflows in future. %

Our framework assumes that the referenced code repositories are reasonably well-structured, up-to-date, and documented. In practice, however, open-source repositories may be poorly documented or incomplete, making them challenging to install autonomously. 
In fact, there is no guarantee that any given repository will be installable and usable as a tool.
For \ourbenchmark, we manually curated the tasks such that we were able to successfully install and use the repository ourselves.
This way, we ensured the tasks were \emph{possible} in the first place.

While \ourbenchmark contains over 100 unit tests, passing these tests does not guarantee correctness in all real-world scenarios. Scientific workflows often involve edge cases or unexpected patterns that are not captured by a small set of tests. Moreover, high-stakes applications such as clinical research would naturally demand additional layers of rigorous validation and oversight by domain experts.

Finally, while \ourbenchmark pins the exact commits of the referenced repositories, external factors such as repository deletion, force-pushing changes, or renaming branches, could affect reproducibility.

\ifdefined\isanonymous\else
\vspace{.5em}
\small{
\noindent\textbf{Acknowledgments}
We thank Junhao Liang, Michaela Unger, and David Charatan for contributing tasks to \ourbenchmark.
We also appreciate Jan Clusmann, Tim Lenz, and Lina Hadji-Kyriacou for their feedback on the manuscript, 
and thank Nathaly Dongo and Annelies Bl\"{a}tterlein for logo design.

\noindent\textbf{Funding}
GW is supported by SCADS.AI, Lothian NHS, and in part by funding from the European Union's Horizon 2020 research and innovation programme (KATY, 101017453). 
JNK is supported by the German Cancer Aid (DECADE, 70115166), the German Federal Ministry of Education and Research (PEARL, 01KD2104C; CAMINO, 01EO2101; TRANSFORM LIVER, 031L0312A; TANGERINE, 01KT2302 through ERA-NET Transcan; Come2Data, 16DKZ2044A; DEEP-HCC, 031L0315A; DECIPHER-M, 01KD2420A; NextBIG, 01ZU2402A), the German Academic Exchange Service (SECAI, 57616814), the German Federal Joint Committee (TransplantKI, 01VSF21048), the European Union's Horizon Europe research and innovation programme (ODELIA, 101057091; GENIAL, 101096312), the European Research Council (ERC; NADIR, 101114631), the National Institutes of Health (EPICO, R01 CA263318) and the National Institute for Health and Care Research (NIHR) Leeds Biomedical Research Centre (grant number NIHR203331). %
}
\fi

\bibliography{bibliography,tool_papers}

\newpage
\appendix
\onecolumn
\section{\Ours}

\subsection{Detailed workflow description}
We provide a detailed description of every step in the \ours workflow to supplement \cref{alg:toolmaker} and our discussion thereof in \cref{sec:toolmaker_workflow}.

\subsubsection{Setting up the environment}
The environment definition is a state of the world (\eg the operating system) that is required for the tool created by \ours to execute.
We can represent this state as a sequence of actions (\eg bash commands or instructions in a Dockerfile, as shown in \cref{fig:tool_creation}, left) that mutate a known initial state (\eg a freshly installed operating system) to the state required for the tool to execute.

To obtain the state of the execution environment necessary for the tool to execute, we employ an agent\:\icon{images/icons/agent.pdf} that is instructed to install and set up the repository (we provide the full prompt in \cref{app:prompts}).
This agent will clone and explore the repository, read documentation, and download any dependencies it deems necessary such as models, datasets, and libraries. Each of these steps involve planning and learning from previous mistakes such as error logs arising during execution.
The agent begins with a clean state (a \mbox{\texttt{python:3.12}} Docker image).
Importantly, we record all actions\:\icon{images/icons/environment_interaction.pdf} that the agent performs.
Since each of the write actions can be expressed as a bash command, we can simply concatenate the bash representations of all recorded write actions to obtain the environment definition in the form of a bash script or Dockerfile.

\subsubsection{Initial tool implementation}
Equipped with the environment definition, which allows \ours to reset the state of the execution environment to the state in which the tool should be executed, it can now implement the tool itself.
Note that we do not carry over the conversation history from the previous stage, in order to not pollute the context window with a large number of messages that are irrelevant for this stage.

\paragraph{\icon{images/icons/agent.pdf}\:Gather information}
We first instruct an agent to explore the installed repository and gather all information necessary to implement the tool.
We include the \emph{tool definition} (see \cref{fig:overview}, top left) as a Python function signature with a docstring in the initial prompt, so that it can use the information it has already gathered to create the plan.

\paragraph{\icon{images/icons/llm_call.pdf}\:Create a plan}
Then, we perform an \gls{llm} call to create a step-by-step plan for the tool implementation.
Here, we keep all of the agent's messages (including actions and observations) in the conversation history, so that it can use the information it has already gathered to create the plan.

\paragraph{\icon{images/icons/llm_call.pdf}\:Implement the tool function}
Next, we instruct the \gls{llm} to implement the tool based on the plan.
Again, we keep the entire conversation history in the context window of the \gls{llm} call, so that it can refer to previous messages.
We now have our first \emph{candidate implementation} of the tool function.

We use OpenAI's \texttt{o1-mini-2024-09-12} model for the planning step as well as the implementation step, to take advantage of its reasoning and code generation capabilities.

\subsubsection{Closed-loop self-improvement}
\label{app:toolmaker_workflow:iteratively_fixing}

\paragraph{\icon{images/icons/environment_interaction.pdf}\:Run the tool}
Before executing the candidate implementation, we \emph{reset} the execution environment to the \emph{environment definition} because the agent may have performed write actions in the past (either in the process of exploring the repository, or in a previous iteration of the loop).
Then, we run the candidate Python function in the execution environment, using the example invocation provided in the tool definition.

\paragraph{\icon{images/icons/llm_call.pdf}\:Assess tool execution}
We instruct the \gls{llm} to assess whether the execution was successful, based on the result returned by the function, as well as the standard output and standard error streams produced during function execution.
Specifically, we ask the \gls{llm} to check whether the result returned by the tool is in line with the task description (\ie if the result is plausible), and whether the standard output and standard error streams contain any indications of errors.
If the \gls{llm} determines that the tool execution was successful, we have arrived at our final tool implementation, and exit the loop.
Otherwise, we continue the self-improvement loop.

\paragraph{\icon{images/icons/agent.pdf}\:Diagnose error}
We instruct an agent to gather information about the error in order to diagnose its root cause, and to formulate a plan to fix the error.
Importantly, we do not reset the execution environment -- the agent is able to check intermediate files and outputs created during tool execution.

\paragraph{\icon{images/icons/llm_call.pdf}\:Re-implement the tool function}
We perform an \gls{llm} call to re-implement the tool based on the current implementation, the error diagnosis, and the plan to fix the error.

\paragraph{\icon{images/icons/llm_call.pdf}\:Summarise the attempt}
Given the conversation history of the current attempt, we instruct the \gls{llm} to summarise the attempt (\ie the diagnosed error and steps taken to fix the error).

This concludes the current attempt.
We reset the state of the execution environment to the environment definition.
We also reset the conversation history to the state before the current attempt started (\ie immediately after the initial implementation of the tool function). However, we append the summaries of all past attempts including the current one to the conversation history, and also include the current version of the code implementation.
Then, we continue with the next iteration of the loop, \ie go back to the start of \cref{app:toolmaker_workflow:iteratively_fixing}.

\section{Extended results}
\label{app:extended_results}

\subsection{Per-task ablation results}
\label{app:extended_ablation_results}
In \cref{tab:results_with_paper,tab:results_with_o3mini,tab:results_with_claude35sonnet}, we provide detailed extended results for the ablations in a format similar to \cref{tab:results} in the main paper.

\begin{table*}
  \centering
  \adjustbox{max width=\linewidth}{
\begin{tabular}{ll|rrrrr|rrrrr}
\toprule
&  & \multicolumn{5}{c}{\textbf{\ours (ours)}} & \multicolumn{5}{c}{\textbf{OpenHands~\cite{wang2024openhands}}} \\
\multicolumn{2}{c|}{\textbf{Task}} & \textbf{Invoc.} & \textbf{Tests} & \textbf{Cost} & \textbf{Actions} & \textbf{Tokens} & \textbf{Invoc.} & \textbf{Tests} & \textbf{Cost} & \textbf{Actions} & \textbf{Tokens} \\
\midrule
\multirow[t]{6}{*}{Pathology} & \texttt{conch\_extract\_features}~\cite{lu2024conch} & \cellcolor{cellgreen} 3/3 & \cellcolor{cellgreen} 9/9 & \$0.57 & 15 ($2_\circlearrowleft$) & 274,256 & \cellcolor{cellred} \errorinstallfailed & \cellcolor{cellred} \errorinstallfailed & \$0.10 & 6 & 65,957 \\
 & \texttt{musk\_extract\_features}~\cite{xiang2025musk} & \cellcolor{cellred} 0/3 & \cellcolor{cellyellow} 3/6 & \$0.68 & 19 ($3_\circlearrowleft$) & 355,561 & \cellcolor{cellred} \errorinstallfailed & \cellcolor{cellred} \errorinstallfailed & \$0.12 & 6 & 77,416 \\
 & \texttt{pathfinder\_verify\_biomarker}~\cite{liang2023pathfinder} & \cellcolor{cellred} 0/2 & \cellcolor{cellyellow} 4/6 & \$0.75 & 25 ($1_\circlearrowleft$) & 473,741 & \cellcolor{cellred} 0/2 & \cellcolor{cellyellow} 4/6 & \$0.11 & 7 & 69,545 \\
 & \texttt{stamp\_extract\_features}~\cite{elnahhas2024stamp} & \cellcolor{cellgreen} 3/3 & \cellcolor{cellgreen} 12/12 & \$1.13 & 20 ($4_\circlearrowleft$) & 649,284 & \cellcolor{cellred} 0/3 & \cellcolor{cellyellow} 3/12 & \$0.08 & 7 & 52,596 \\
 & \texttt{stamp\_train\_classification\_model}~\cite{elnahhas2024stamp} & \cellcolor{cellgreen} 3/3 & \cellcolor{cellgreen} 9/9 & \$0.76 & 15 ($5_\circlearrowleft$) & 393,150 & \cellcolor{cellred} 0/3 & \cellcolor{cellred} 0/9 & \$0.25 & 11 & 143,934 \\
 & \texttt{uni\_extract\_features}~\cite{chen2024uni} & \cellcolor{cellgreen} 3/3 & \cellcolor{cellgreen} 9/9 & \$0.53 & 14 ($3_\circlearrowleft$) & 268,481 & \cellcolor{cellgreen} 3/3 & \cellcolor{cellgreen} 9/9 & \$0.14 & 5 & 87,344 \\
\hline
\bottomrule
\end{tabular}

  }
  \caption{Results (with paper summary in context).}
  \label{tab:results_with_paper}
\end{table*}
\begin{table*}
  \centering
  \adjustbox{max width=\linewidth}{
    \begin{tabular}{ll|rrrrr|rrrrr}
\toprule
&  & \multicolumn{5}{c}{\textbf{\ours (ours)}} & \multicolumn{5}{c}{\textbf{OpenHands~\cite{wang2024openhands}}} \\
\multicolumn{2}{c|}{\textbf{Task}} & \textbf{Invoc.} & \textbf{Tests} & \textbf{Cost} & \textbf{Actions} & \textbf{Tokens} & \textbf{Invoc.} & \textbf{Tests} & \textbf{Cost} & \textbf{Actions} & \textbf{Tokens} \\
\midrule
\multirow[t]{6}{*}{Pathology} & \texttt{conch\_extract\_features}~\cite{lu2024conch} & \cellcolor{cellred} 0/3 & \cellcolor{cellyellow} 6/9 & \$0.22 & 15 ($2_\circlearrowleft$) & 232,441 & \cellcolor{cellred} \errorinstallfailed & \cellcolor{cellred} \errorinstallfailed & \$0.04 & 0 & 28,880 \\
 & \texttt{musk\_extract\_features}~\cite{xiang2025musk} & \cellcolor{cellred} 0/3 & \cellcolor{cellyellow} 3/6 & \$0.24 & 18 ($2_\circlearrowleft$) & 247,840 & \cellcolor{cellred} \errorinstallfailed & \cellcolor{cellred} \errorinstallfailed & \$0.03 & 2 & 22,820 \\
 & \texttt{pathfinder\_verify\_biomarker}~\cite{liang2023pathfinder} & \cellcolor{cellred} 0/2 & \cellcolor{cellyellow} 4/6 & \$0.10 & 11 ($1_\circlearrowleft$) & 85,312 & \cellcolor{cellred} \errorinstallfailed & \cellcolor{cellred} \errorinstallfailed & \$0.04 & 2 & 25,797 \\
 & \texttt{stamp\_extract\_features}~\cite{elnahhas2024stamp} & \cellcolor{cellgreen} 3/3 & \cellcolor{cellgreen} 12/12 & \$0.18 & 14 ($1_\circlearrowleft$) & 187,972 & \cellcolor{cellred} \errorinstallfailed & \cellcolor{cellred} \errorinstallfailed & \$0.04 & 2 & 23,343 \\
 & \texttt{stamp\_train\_classification\_model}~\cite{elnahhas2024stamp} & \cellcolor{cellgreen} 3/3 & \cellcolor{cellgreen} 9/9 & \$0.38 & 17 ($4_\circlearrowleft$) & 403,138 & \cellcolor{cellred} 0/3 & \cellcolor{cellred} 0/9 & \$0.04 & 2 & 32,516 \\
 & \texttt{uni\_extract\_features}~\cite{chen2024uni} & \cellcolor{cellgreen} 3/3 & \cellcolor{cellgreen} 9/9 & \$0.58 & 12 ($8_\circlearrowleft$) & 563,488 & \cellcolor{cellred} 0/3 & \cellcolor{cellred} 0/9 & \$0.03 & 2 & 22,905 \\
\hline
\bottomrule
\end{tabular}

  }
  \caption{Results (using o3-mini).}
  \label{tab:results_with_o3mini}
\end{table*}
\begin{table*}
  \centering
  \adjustbox{max width=\linewidth}{
    \begin{tabular}{ll|rrrrr}
\toprule
&  & \multicolumn{5}{c}{\textbf{OpenHands~\cite{wang2024openhands}}} \\
\multicolumn{2}{c|}{\textbf{Task}} & \textbf{Invoc.} & \textbf{Tests} & \textbf{Cost} & \textbf{Actions} & \textbf{Tokens} \\
\midrule
\multirow[t]{6}{*}{Pathology} & \texttt{conch\_extract\_features}~\cite{lu2024conch} & \cellcolor{cellgreen} 3/3 & \cellcolor{cellgreen} 9/9 & \$0.12 & 4 & 59,911 \\
 & \texttt{musk\_extract\_features}~\cite{xiang2025musk} & \cellcolor{cellred} 0/3 & \cellcolor{cellred} 0/6 & \$0.09 & 3 & 45,985 \\
 & \texttt{pathfinder\_verify\_biomarker}~\cite{liang2023pathfinder} & \cellcolor{cellred} 0/2 & \cellcolor{cellyellow} 4/6 & \$0.09 & 4 & 49,661 \\
 & \texttt{stamp\_extract\_features}~\cite{elnahhas2024stamp} & \cellcolor{cellred} \errorinstallfailed & \cellcolor{cellred} \errorinstallfailed & \$0.05 & 2 & 24,799 \\
 & \texttt{stamp\_train\_classification\_model}~\cite{elnahhas2024stamp} & \cellcolor{cellred} 0/3 & \cellcolor{cellred} 0/9 & \$0.13 & 6 & 66,376 \\
 & \texttt{uni\_extract\_features}~\cite{chen2024uni} & \cellcolor{cellred} 0/3 & \cellcolor{cellred} 0/9 & \$0.11 & 8 & 82,516 \\
\hline
\bottomrule
\end{tabular}

  }
  \caption{Results (using Claude 3.5 Sonnet).}
  \label{tab:results_with_claude35sonnet}
\end{table*}

\subsection{Raw unit test results}
We provide the raw unit test results for all tasks in \cref{tab:raw_toolmaker,tab:raw_openhands} for the main experiments and \cref{tab:raw_toolmaker_paper,tab:raw_openhands_paper,tab:raw_toolmaker_o3mini,tab:raw_openhands_o3mini,tab:raw_openhands_claude35sonnet} for the ablations.
\begin{table}
  \centering
  \begin{adjustbox}{max totalheight=.9\textheight}
  \begin{tabular}{lllll}
\toprule
 &  &  &  & Passed \\
Category & Task & Call & Test &  \\
\midrule
\multirow[t]{51}{*}{Pathology} & \multirow[t]{9}{*}{\texttt{conch\_extract\_features}~\cite{lu2024conch}} & \multirow[t]{3}{*}{\texttt{kather100k\_muc}} & \texttt{test\_feature\_values} & \cmark \\
 &  &  & \texttt{test\_shape\_and\_type} & \cmark \\
 &  &  & \texttt{test\_status} & \cmark \\
\cline{3-5}
 &  & \multirow[t]{3}{*}{\texttt{tcga\_brca\_patch\_jpg}} & \texttt{test\_feature\_values} & \cmark \\
 &  &  & \texttt{test\_shape\_and\_type} & \cmark \\
 &  &  & \texttt{test\_status} & \cmark \\
\bottomrule
\end{tabular}

  \end{adjustbox}
  \caption{Raw results (\ours, without paper summary in context).}
  \label{tab:raw_toolmaker}
\end{table}
\begin{table}
  \centering
  \begin{adjustbox}{max totalheight=.9\textheight}
  \begin{tabular}{lllll}
\toprule
 &  &  &  & Passed \\
Category & Task & Call & Test &  \\
\midrule
\multirow[t]{51}{*}{Pathology} & \multirow[t]{9}{*}{\texttt{conch\_extract\_features}~\cite{lu2024conch}} & \multirow[t]{3}{*}{\texttt{kather100k\_muc}} & \texttt{test\_feature\_values} & \cmark \\
 &  &  & \texttt{test\_shape\_and\_type} & \cmark \\
 &  &  & \texttt{test\_status} & \cmark \\
\cline{3-5}
 &  & \multirow[t]{3}{*}{\texttt{tcga\_brca\_patch\_jpg}} & \texttt{test\_feature\_values} & \cmark \\
 &  &  & \texttt{test\_shape\_and\_type} & \cmark \\
 &  &  & \texttt{test\_status} & \cmark \\
\bottomrule
\end{tabular}

  \end{adjustbox}
  \caption{Raw results (OpenHands~\cite{wang2024openhands}, without paper summary in context).}
  \label{tab:raw_openhands}
\end{table}
\begin{table}
  \centering
  \begin{adjustbox}{max totalheight=.9\textheight}
  \begin{tabular}{lllll}
\toprule
 &  &  &  & Passed \\
Category & Task & Call & Test &  \\
\midrule
\multirow[t]{51}{*}{Pathology} & \multirow[t]{9}{*}{\texttt{conch\_extract\_features}~\cite{lu2024conch}} & \multirow[t]{3}{*}{\texttt{kather100k\_muc}} & \texttt{test\_feature\_values} & \cmark \\
 &  &  & \texttt{test\_shape\_and\_type} & \cmark \\
 &  &  & \texttt{test\_status} & \cmark \\
\cline{3-5}
 &  & \multirow[t]{3}{*}{\texttt{tcga\_brca\_patch\_jpg}} & \texttt{test\_feature\_values} & \cmark \\
 &  &  & \texttt{test\_shape\_and\_type} & \cmark \\
 &  &  & \texttt{test\_status} & \cmark \\
\bottomrule
\end{tabular}

  \end{adjustbox}
  \caption{Raw results (\ours, with paper summary in context).}
  \label{tab:raw_toolmaker_paper}
\end{table}
\begin{table}
  \centering
  \begin{adjustbox}{max totalheight=.9\textheight}
  \begin{tabular}{lllll}
\toprule
 &  &  &  & Passed \\
Category & Task & Call & Test &  \\
\midrule
\multirow[t]{51}{*}{Pathology} & \multirow[t]{9}{*}{\texttt{conch\_extract\_features}~\cite{lu2024conch}} & \multirow[t]{3}{*}{\texttt{kather100k\_muc}} & \texttt{test\_feature\_values} & \xmark \\
 &  &  & \texttt{test\_shape\_and\_type} & \xmark \\
 &  &  & \texttt{test\_status} & \xmark \\
\cline{3-5}
 &  & \multirow[t]{3}{*}{\texttt{tcga\_brca\_patch\_jpg}} & \texttt{test\_feature\_values} & \xmark \\
 &  &  & \texttt{test\_shape\_and\_type} & \xmark \\
 &  &  & \texttt{test\_status} & \xmark \\
\bottomrule
\end{tabular}

  \end{adjustbox}
  \caption{Raw results (OpenHands~\cite{wang2024openhands}, with paper summary in context).}
  \label{tab:raw_openhands_paper}
\end{table}
\begin{table}
  \centering
  \begin{adjustbox}{max totalheight=.9\textheight}
  \begin{tabular}{lllll}
\toprule
 &  &  &  & Passed \\
Category & Task & Call & Test &  \\
\midrule
\multirow[t]{51}{*}{Pathology} & \multirow[t]{9}{*}{\texttt{conch\_extract\_features}~\cite{lu2024conch}} & \multirow[t]{3}{*}{\texttt{kather100k\_muc}} & \texttt{test\_feature\_values} & \xmark \\
 &  &  & \texttt{test\_shape\_and\_type} & \cmark \\
 &  &  & \texttt{test\_status} & \cmark \\
\cline{3-5}
 &  & \multirow[t]{3}{*}{\texttt{tcga\_brca\_patch\_jpg}} & \texttt{test\_feature\_values} & \xmark \\
 &  &  & \texttt{test\_shape\_and\_type} & \cmark \\
 &  &  & \texttt{test\_status} & \cmark \\
\bottomrule
\end{tabular}

  \end{adjustbox}
  \caption{Raw results (\ours, using o3-mini instead of gpt-4o).}
  \label{tab:raw_toolmaker_o3mini}
\end{table}
\begin{table}
  \centering
  \begin{adjustbox}{max totalheight=.9\textheight}
  \begin{tabular}{lllll}
\toprule
 &  &  &  & Passed \\
Category & Task & Call & Test &  \\
\midrule
\multirow[t]{51}{*}{Pathology} & \multirow[t]{9}{*}{\texttt{conch\_extract\_features}~\cite{lu2024conch}} & \multirow[t]{3}{*}{\texttt{kather100k\_muc}} & \texttt{test\_feature\_values} & \xmark \\
 &  &  & \texttt{test\_shape\_and\_type} & \xmark \\
 &  &  & \texttt{test\_status} & \xmark \\
\cline{3-5}
 &  & \multirow[t]{3}{*}{\texttt{tcga\_brca\_patch\_jpg}} & \texttt{test\_feature\_values} & \xmark \\
 &  &  & \texttt{test\_shape\_and\_type} & \xmark \\
 &  &  & \texttt{test\_status} & \xmark \\
\bottomrule
\end{tabular}

  \end{adjustbox}
  \caption{Raw results (OpenHands~\cite{wang2024openhands}, using o3-mini instead of gpt-4o).}
  \label{tab:raw_openhands_o3mini}
\end{table}
\begin{table}
  \centering
  \begin{adjustbox}{max totalheight=.9\textheight}
  \begin{tabular}{lllll}
\toprule
 &  &  &  & Passed \\
Category & Task & Call & Test &  \\
\midrule
\multirow[t]{51}{*}{Pathology} & \multirow[t]{9}{*}{\texttt{conch\_extract\_features}~\cite{lu2024conch}} & \multirow[t]{3}{*}{\texttt{kather100k\_muc}} & \texttt{test\_feature\_values} & \cmark \\
 &  &  & \texttt{test\_shape\_and\_type} & \cmark \\
 &  &  & \texttt{test\_status} & \cmark \\
\cline{3-5}
 &  & \multirow[t]{3}{*}{\texttt{tcga\_brca\_patch\_jpg}} & \texttt{test\_feature\_values} & \cmark \\
 &  &  & \texttt{test\_shape\_and\_type} & \cmark \\
 &  &  & \texttt{test\_status} & \cmark \\
\bottomrule
\end{tabular}

  \end{adjustbox}
  \caption{Raw results (OpenHands~\cite{wang2024openhands}, using Claude 3.5 Sonnet instead of gpt-4o).}
  \label{tab:raw_openhands_claude35sonnet}
\end{table}

\subsection{Transitions between tool calls}
\label{app:transitions}
\begin{figure}
    \centering
    \includegraphics[width=0.8\linewidth]{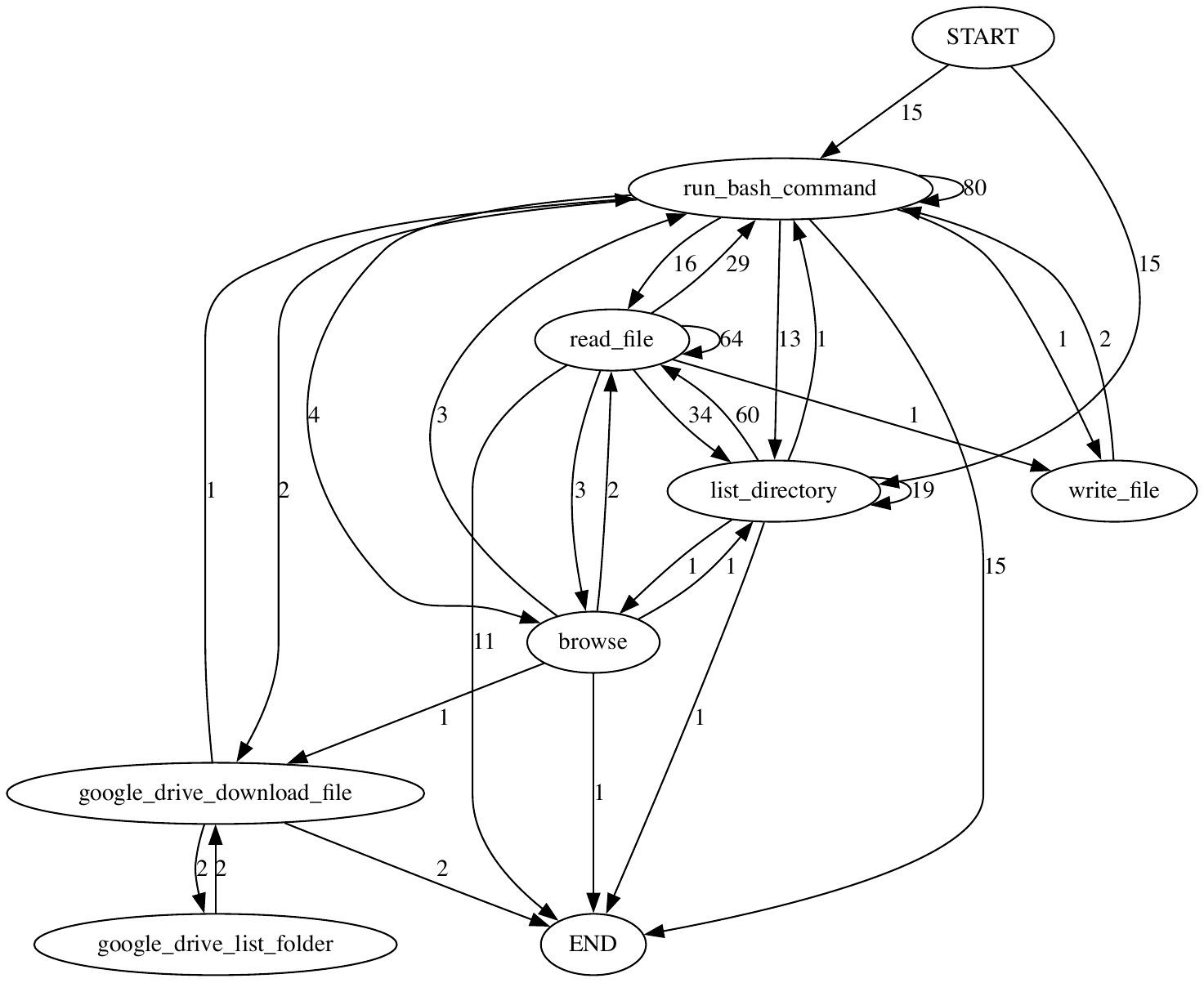}
    \caption{Transitions between tool calls by \ours.}
    \label{fig:trace}
\end{figure}

In \cref{fig:trace}, we show the transitions between tool calls by \ours.

\section{\Ourbenchmark}
\label{app:benchmark}
Below are the complete task definitions for all tasks in our benchmark, \ourbenchmark.

\subsection{Pathology}
\label{app:tasks:pathology}
\begin{tcolorbox}[title={\texttt{conch\_extract\_features}}]
Perform feature extraction on an input image using CONCH.

\vspace{.5em}
\textbf{Arguments:}
\begin{itemize}[topsep=0pt,parsep=-1pt,partopsep=0pt]
\item \texttt{input\_image} (\texttt{str}): Path to the input image\\  Example: \texttt{'/mount/input/TUM/TUM-TCGA-ACRLPPQE.tif'}
\end{itemize}

\vspace{.5em}
\textbf{Returns:} \begin{itemize}[topsep=0pt,parsep=-1pt,partopsep=0pt]
\item \texttt{features} (\texttt{list}): The feature vector extracted from the input image, as a list of floats
\end{itemize}
\tcblower
\setlength{\hangindent}{\widthof{\faGithub~}}
\faGithub~\url{https://github.com/mahmoodlab/CONCH}

\vspace{.5em}\setlength{\hangindent}{\widthof{\faFile*[regular]~}}\faFile*[regular]~\bibentry{lu2024conch}

\end{tcolorbox}

\begin{tcolorbox}[title={\texttt{musk\_extract\_features}}]
Perform feature extraction on an input image using the vision part of MUSK.

\vspace{.5em}
\textbf{Arguments:}
\begin{itemize}[topsep=0pt,parsep=-1pt,partopsep=0pt]
\item \texttt{input\_image} (\texttt{str}): Path to the input image\\  Example: \texttt{'/mount/input/TUM/TUM-TCGA-ACRLPPQE.tif'}
\end{itemize}

\vspace{.5em}
\textbf{Returns:} \begin{itemize}[topsep=0pt,parsep=-1pt,partopsep=0pt]
\item \texttt{features} (\texttt{list}): The feature vector extracted from the input image, as a list of floats
\end{itemize}
\tcblower
\setlength{\hangindent}{\widthof{\faGithub~}}
\faGithub~\url{https://github.com/lilab-stanford/MUSK}

\vspace{.5em}\setlength{\hangindent}{\widthof{\faFile*[regular]~}}\faFile*[regular]~\bibentry{xiang2025musk}

\end{tcolorbox}

\begin{tcolorbox}[title={\texttt{pathfinder\_verify\_biomarker}}]
Given WSI probability maps, a hypothesis of a potential biomarker, and clinical data, determine (1) whether the potential biomarker is significant for patient prognosis, and (2) whether the potential biomarker is independent among already known biomarkers.

\vspace{.5em}
\textbf{Arguments:}
\begin{itemize}[topsep=0pt,parsep=-1pt,partopsep=0pt]
\item \texttt{heatmaps} (\texttt{str}): Path to the folder containing the numpy array (\textasciigrave{}*.npy\textasciigrave{}) files, which contains the heatmaps of the trained model (each heatmap is HxWxC where C is the number of classes)\\  Example: \texttt{'/mount/input/TCGA\_CRC'}
\item \texttt{hypothesis} (\texttt{str}): A python file, which contains a function \textasciigrave{}def hypothesis\_score(prob\_map\_path: str) -\textgreater{} float\textasciigrave{} which expresses a mathematical model of a hypothesis of a potential biomarker.  For a particular patient, the function returns a risk score.\\  Example: \texttt{'/mount/input/mus\_fraction\_score.py'}
\item \texttt{clini\_table} (\texttt{str}): Path to the CSV file containing the clinical data\\  Example: \texttt{'/mount/input/TCGA\_CRC\_info.csv'}
\item \texttt{files\_table} (\texttt{str}): Path to the CSV file containing the mapping between patient IDs (in the PATIENT column) and heatmap filenames (in the FILENAME column)\\  Example: \texttt{'/mount/input/TCGA\_CRC\_files.csv'}
\item \texttt{survival\_time\_column} (\texttt{str}): The name of the column in the clinical data that contains the survival time\\  Example: \texttt{'OS.time'}
\item \texttt{event\_column} (\texttt{str}): The name of the column in the clinical data that contains the event (e.g. death, recurrence, etc.)\\  Example: \texttt{'vital\_status'}
\item \texttt{known\_biomarkers} (\texttt{list}): A list of known biomarkers. These are column names in the clinical data.\\  Example: \texttt{['MSI']}
\end{itemize}

\vspace{.5em}
\textbf{Returns:} \begin{itemize}[topsep=0pt,parsep=-1pt,partopsep=0pt]
\item \texttt{p\_value} (\texttt{float}): The p-value of the significance of the potential biomarker
\item \texttt{hazard\_ratio} (\texttt{float}): The hazard ratio for the biomarker
\end{itemize}
\tcblower
\setlength{\hangindent}{\widthof{\faGithub~}}
\faGithub~\url{https://github.com/LiangJunhao-THU/PathFinderCRC}

\vspace{.5em}\setlength{\hangindent}{\widthof{\faFile*[regular]~}}\faFile*[regular]~\bibentry{liang2023pathfinder}

\end{tcolorbox}

\begin{tcolorbox}[title={\texttt{stamp\_extract\_features}}]
Perform feature extraction using CTransPath with STAMP on a set of whole slide images, and save the resulting features to a new folder.

\vspace{.5em}
\textbf{Arguments:}
\begin{itemize}[topsep=0pt,parsep=-1pt,partopsep=0pt]
\item \texttt{output\_dir} (\texttt{str}): Path to the output folder where the features will be saved\\  Example: \texttt{'/mount/output/TCGA-BRCA-features'}
\item \texttt{slide\_dir} (\texttt{str}): Path to the input folder containing the whole slide images\\  Example: \texttt{'/mount/input/TCGA-BRCA-SLIDES'}
\end{itemize}

\vspace{.5em}
\textbf{Returns:} \begin{itemize}[topsep=0pt,parsep=-1pt,partopsep=0pt]
\item \texttt{num\_processed\_slides} (\texttt{int}): The number of slides that were processed
\end{itemize}
\tcblower
\setlength{\hangindent}{\widthof{\faGithub~}}
\faGithub~\url{https://github.com/KatherLab/STAMP}

\vspace{.5em}\setlength{\hangindent}{\widthof{\faFile*[regular]~}}\faFile*[regular]~\bibentry{elnahhas2024stamp}

\end{tcolorbox}

\begin{tcolorbox}[title={\texttt{stamp\_train\_classification\_model}}]
Train a model for biomarker classification. You will be supplied with the path to the folder containing the whole slide images, alongside a path to a CSV file containing the training labels.

\vspace{.5em}
\textbf{Arguments:}
\begin{itemize}[topsep=0pt,parsep=-1pt,partopsep=0pt]
\item \texttt{slide\_dir} (\texttt{str}): Path to the folder containing the whole slide images\\  Example: \texttt{'/mount/input/TCGA-BRCA-SLIDES'}
\item \texttt{clini\_table} (\texttt{str}): Path to the CSV file containing the clinical data\\  Example: \texttt{'/mount/input/TCGA-BRCA-DX\_CLINI.xlsx'}
\item \texttt{slide\_table} (\texttt{str}): Path to the CSV file containing the slide metadata\\  Example: \texttt{'/mount/input/TCGA-BRCA-DX\_SLIDE.csv'}
\item \texttt{target\_column} (\texttt{str}): The name of the column in the clinical data that contains the target labels\\  Example: \texttt{'TP53\_driver'}
\item \texttt{trained\_model\_path} (\texttt{str}): Path to the *.pkl file where the trained model should be saved by this function\\  Example: \texttt{'/mount/output/STAMP-BRCA-TP53-model.pkl'}
\end{itemize}

\vspace{.5em}
\textbf{Returns:} \begin{itemize}[topsep=0pt,parsep=-1pt,partopsep=0pt]
\item \texttt{num\_params} (\texttt{int}): The number of parameters in the trained model
\end{itemize}
\tcblower
\setlength{\hangindent}{\widthof{\faGithub~}}
\faGithub~\url{https://github.com/KatherLab/STAMP}

\vspace{.5em}\setlength{\hangindent}{\widthof{\faFile*[regular]~}}\faFile*[regular]~\bibentry{elnahhas2024stamp}

\end{tcolorbox}

\begin{tcolorbox}[title={\texttt{uni\_extract\_features}}]
Perform feature extraction on an input image using UNI.

\vspace{.5em}
\textbf{Arguments:}
\begin{itemize}[topsep=0pt,parsep=-1pt,partopsep=0pt]
\item \texttt{input\_image} (\texttt{str}): Path to the input image\\  Example: \texttt{'/mount/input/TUM/TUM-TCGA-ACRLPPQE.tif'}
\end{itemize}

\vspace{.5em}
\textbf{Returns:} \begin{itemize}[topsep=0pt,parsep=-1pt,partopsep=0pt]
\item \texttt{features} (\texttt{list}): The feature vector extracted from the input image, as a list of floats
\end{itemize}
\tcblower
\setlength{\hangindent}{\widthof{\faGithub~}}
\faGithub~\url{https://github.com/mahmoodlab/UNI}

\vspace{.5em}\setlength{\hangindent}{\widthof{\faFile*[regular]~}}\faFile*[regular]~\bibentry{chen2024uni}

\end{tcolorbox}

\subsection{Radiology}
\label{app:tasks:radiology}
\begin{tcolorbox}[title={\texttt{medsam\_inference}}]
Use the trained MedSAM model to segment the given abdomen CT scan.

\vspace{.5em}
\textbf{Arguments:}
\begin{itemize}[topsep=0pt,parsep=-1pt,partopsep=0pt]
\item \texttt{image\_file} (\texttt{str}): Path to the abdomen CT scan image.\\  Example: \texttt{'/mount/input/my\_image.jpg'}
\item \texttt{bounding\_box} (\texttt{list}): Bounding box to segment (list of 4 integers).\\  Example: \texttt{[25, 100, 155, 155]}
\item \texttt{segmentation\_file} (\texttt{str}): Path to where the segmentation image should be saved.\\  Example: \texttt{'/mount/output/segmented\_image.png'}
\end{itemize}

\vspace{.5em}
\textbf{Returns:} \textit{empty dict}
\tcblower
\setlength{\hangindent}{\widthof{\faGithub~}}
\faGithub~\url{https://github.com/bowang-lab/MedSAM}

\vspace{.5em}\setlength{\hangindent}{\widthof{\faFile*[regular]~}}\faFile*[regular]~\bibentry{ma2024medsam}

\end{tcolorbox}

\begin{tcolorbox}[title={\texttt{nnunet\_train\_model}}]
Train a nnUNet model from scratch on abdomen CT scans. You will be provided with  the path to the dataset, the nnUNet configuration to use, and the fold number  to train the model on.

\vspace{.5em}
\textbf{Arguments:}
\begin{itemize}[topsep=0pt,parsep=-1pt,partopsep=0pt]
\item \texttt{dataset\_path} (\texttt{str}): The path to the dataset to train the model on (contains dataset.json, imagesTr, imagesTs, labelsTr)\\  Example: \texttt{'/mount/input/Task02\_Heart'}
\item \texttt{unet\_configuration} (\texttt{str}): The configuration of the UNet to use for training. One of '2d', '3d\_fullres', '3d\_lowres', '3d\_cascade\_fullres'\\  Example: \texttt{'3d\_fullres'}
\item \texttt{fold} (\texttt{int}): The fold number to train the model on. One of 0, 1, 2, 3, 4.\\  Example: \texttt{0}
\item \texttt{output\_folder} (\texttt{str}): Path to the folder where the trained model should be saved\\  Example: \texttt{'/mount/output/trained\_model'}
\end{itemize}

\vspace{.5em}
\textbf{Returns:} \textit{empty dict}
\tcblower
\setlength{\hangindent}{\widthof{\faGithub~}}
\faGithub~\url{https://github.com/MIC-DKFZ/nnUNet}

\vspace{.5em}\setlength{\hangindent}{\widthof{\faFile*[regular]~}}\faFile*[regular]~\bibentry{isensee2020nnunet}

\end{tcolorbox}

\subsection{Omics}
\label{app:tasks:genomics_proteomics}
\begin{tcolorbox}[title={\texttt{cytopus\_db}}]
Initialize the Cytopus KnowledgeBase and generate a JSON file containing a nested dictionary with gene set annotations organized by cell type, suitable for input into the Spectra library.

\vspace{.5em}
\textbf{Arguments:}
\begin{itemize}[topsep=0pt,parsep=-1pt,partopsep=0pt]
\item \texttt{celltype\_of\_interest} (\texttt{list}): List of cell types for which to retrieve gene sets\\  Example: \texttt{['B\_memory', 'B\_naive', 'CD4\_T', 'CD8\_T', 'DC', 'ILC3', 'MDC', 'NK', 'Treg', 'gdT', 'mast', 'pDC', 'plasma']}
\item \texttt{global\_celltypes} (\texttt{list}): List of global cell types to include in the JSON file.\\  Example: \texttt{['all-cells', 'leukocyte']}
\item \texttt{output\_file} (\texttt{str}): Path to the file where the output JSON file should be stored.\\  Example: \texttt{'/mount/output/Spectra\_dict.json'}
\end{itemize}

\vspace{.5em}
\textbf{Returns:} \begin{itemize}[topsep=0pt,parsep=-1pt,partopsep=0pt]
\item \texttt{keys} (\texttt{list}): The list of keys in the produced JSON file.
\end{itemize}
\tcblower
\setlength{\hangindent}{\widthof{\faGithub~}}
\faGithub~\url{https://github.com/wallet-maker/cytopus}

\vspace{.5em}\setlength{\hangindent}{\widthof{\faFile*[regular]~}}\faFile*[regular]~\bibentry{kunes2023cytopus}

\end{tcolorbox}

\begin{tcolorbox}[title={\texttt{esm\_fold\_predict}}]
Generate the representation of a protein sequence and the contact map using Facebook Research's pretrained esm2\_t33\_650M\_UR50D model.

\vspace{.5em}
\textbf{Arguments:}
\begin{itemize}[topsep=0pt,parsep=-1pt,partopsep=0pt]
\item \texttt{sequence} (\texttt{str}): Protein sequence to for which to generate representation and contact map.\\  Example: \texttt{'MKTVRQERLKSIVRILERSKEPVSGAQLAEELSVSRQVIVQDIAYLRSLGYNIVATPRGYVLAGG'}
\end{itemize}

\vspace{.5em}
\textbf{Returns:} \begin{itemize}[topsep=0pt,parsep=-1pt,partopsep=0pt]
\item \texttt{sequence\_representation} (\texttt{list}): Token representations for the protein sequence as a list of floats, i.e. a 1D array of shape L where L is the number of tokens.
\item \texttt{contact\_map} (\texttt{list}): Contact map for the protein sequence as a list of list of floats, i.e. a 2D array of shape LxL where L is the number of tokens.
\end{itemize}
\tcblower
\setlength{\hangindent}{\widthof{\faGithub~}}
\faGithub~\url{https://github.com/facebookresearch/esm}

\vspace{.5em}\setlength{\hangindent}{\widthof{\faFile*[regular]~}}\faFile*[regular]~\bibentry{verkuil2022esm1}

\vspace{.5em}\setlength{\hangindent}{\widthof{\faFile*[regular]~}}\faFile*[regular]~\bibentry{hie2022esm2}

\end{tcolorbox}

\subsection{Other}
\label{app:tasks:imaging}
\begin{tcolorbox}[title={\texttt{retfound\_feature\_vector}}]
Extract the feature vector for the given retinal image using the RETFound pretrained vit\_large\_patch16 model.

\vspace{.5em}
\textbf{Arguments:}
\begin{itemize}[topsep=0pt,parsep=-1pt,partopsep=0pt]
\item \texttt{image\_file} (\texttt{str}): Path to the retinal image.\\  Example: \texttt{'/mount/input/retinal\_image.jpg'}
\end{itemize}

\vspace{.5em}
\textbf{Returns:} \begin{itemize}[topsep=0pt,parsep=-1pt,partopsep=0pt]
\item \texttt{feature\_vector} (\texttt{list}): The feature vector for the given retinal image, as a list of floats.
\end{itemize}
\tcblower
\setlength{\hangindent}{\widthof{\faGithub~}}
\faGithub~\url{https://github.com/rmaphoh/RETFound_MAE}

\vspace{.5em}\setlength{\hangindent}{\widthof{\faFile*[regular]~}}\faFile*[regular]~\bibentry{zhou2023retfound}

\end{tcolorbox}

\label{app:tasks:llms}
\begin{tcolorbox}[title={\texttt{medsss\_generate}}]
Given a user message, generate a response using the MedSSS\_Policy model.

\vspace{.5em}
\textbf{Arguments:}
\begin{itemize}[topsep=0pt,parsep=-1pt,partopsep=0pt]
\item \texttt{user\_message} (\texttt{str}): The user message.\\  Example: \texttt{'How to stop a cough?'}
\end{itemize}

\vspace{.5em}
\textbf{Returns:} \begin{itemize}[topsep=0pt,parsep=-1pt,partopsep=0pt]
\item \texttt{response} (\texttt{str}): The response generated by the model.
\end{itemize}
\tcblower
\setlength{\hangindent}{\widthof{\faGithub~}}
\faGithub~\url{https://github.com/pixas/MedSSS}

\vspace{.5em}\setlength{\hangindent}{\widthof{\faFile*[regular]~}}\faFile*[regular]~\bibentry{jiang2025medsss}

\end{tcolorbox}

\begin{tcolorbox}[title={\texttt{modernbert\_predict\_masked}}]
Given a masked sentence string, predict the original sentence using the pretrained ModernBERT-base model on CPU.

\vspace{.5em}
\textbf{Arguments:}
\begin{itemize}[topsep=0pt,parsep=-1pt,partopsep=0pt]
\item \texttt{input\_string} (\texttt{str}): The masked sentence string. The masked part is represented by "[MASK]"".\\  Example: \texttt{'Paris is the [MASK] of France.'}
\end{itemize}

\vspace{.5em}
\textbf{Returns:} \begin{itemize}[topsep=0pt,parsep=-1pt,partopsep=0pt]
\item \texttt{prediction} (\texttt{str}): The predicted original sentence
\end{itemize}
\tcblower
\setlength{\hangindent}{\widthof{\faGithub~}}
\faGithub~\url{https://github.com/AnswerDotAI/ModernBERT}

\vspace{.5em}\setlength{\hangindent}{\widthof{\faFile*[regular]~}}\faFile*[regular]~\bibentry{warner2024modernbert}

\end{tcolorbox}

\label{app:tasks:3d_vision}
\begin{tcolorbox}[title={\texttt{flowmap\_overfit\_scene}}]
Overfit FlowMap on an input scene to determine camera extrinsics for each frame in the scene.

\vspace{.5em}
\textbf{Arguments:}
\begin{itemize}[topsep=0pt,parsep=-1pt,partopsep=0pt]
\item \texttt{input\_scene} (\texttt{str}): Path to the directory containing the images of the input scene (just the image files, nothing else)\\  Example: \texttt{'/mount/input/llff\_flower'}
\end{itemize}

\vspace{.5em}
\textbf{Returns:} \begin{itemize}[topsep=0pt,parsep=-1pt,partopsep=0pt]
\item \texttt{n} (\texttt{int}): The number of images (frames) in the scene
\item \texttt{camera\_extrinsics} (\texttt{list}): The camera extrinsics matrix for each of the n frames in the scene, must have a shape of nx4x4 (as a nested python list of floats)
\end{itemize}
\tcblower
\setlength{\hangindent}{\widthof{\faGithub~}}
\faGithub~\url{https://github.com/dcharatan/flowmap}

\vspace{.5em}\setlength{\hangindent}{\widthof{\faFile*[regular]~}}\faFile*[regular]~\bibentry{smith2024flowmap}

\end{tcolorbox}

\label{app:tasks:tabular}
\begin{tcolorbox}[title={\texttt{tabpfn\_predict}}]
Train a predictor using TabPFN on a tabular dataset. Evaluate the predictor on the test set.

\vspace{.5em}
\textbf{Arguments:}
\begin{itemize}[topsep=0pt,parsep=-1pt,partopsep=0pt]
\item \texttt{train\_csv} (\texttt{str}): Path to the CSV file containing the training data\\  Example: \texttt{'/mount/input/breast\_cancer\_train.csv'}
\item \texttt{test\_csv} (\texttt{str}): Path to the CSV file containing the test data\\  Example: \texttt{'/mount/input/breast\_cancer\_test.csv'}
\item \texttt{feature\_columns} (\texttt{list}): The names of the columns to use as features\\  Example: \texttt{['mean radius', 'mean texture', 'mean perimeter', 'mean area', 'mean smoothness', 'mean compactness', 'mean concavity', 'mean concave points', 'mean symmetry', 'mean fractal dimension', 'radius error', 'texture error', 'perimeter error', 'area error', 'smoothness error', 'compactness error', 'concavity error', 'concave points error', 'symmetry error', 'fractal dimension error', 'worst radius', 'worst texture', 'worst perimeter', 'worst area', 'worst smoothness', 'worst compactness', 'worst concavity', 'worst concave points', 'worst symmetry', 'worst fractal dimension']}
\item \texttt{target\_column} (\texttt{str}): The name of the column to predict\\  Example: \texttt{'target'}
\end{itemize}

\vspace{.5em}
\textbf{Returns:} \begin{itemize}[topsep=0pt,parsep=-1pt,partopsep=0pt]
\item \texttt{roc\_auc} (\texttt{float}): The ROC AUC score of the predictor on the test set
\item \texttt{accuracy} (\texttt{float}): The accuracy of the predictor on the test set
\item \texttt{probs} (\texttt{list}): The probabilities of the predictor on the test set, as a list of floats (one per sample in the test set)
\end{itemize}
\tcblower
\setlength{\hangindent}{\widthof{\faGithub~}}
\faGithub~\url{https://github.com/PriorLabs/TabPFN}

\vspace{.5em}\setlength{\hangindent}{\widthof{\faFile*[regular]~}}\faFile*[regular]~\bibentry{hollmann2025tabpfn}

\end{tcolorbox}

\section{\Ours prompts}
\label{app:prompts}
\begin{tcolorbox}[title={\texttt{Install Repository System Instructions}}]
You're a diligent software engineer AI. You can't see, draw, or interact with
a browser, but you can read and write files, and you can run commands, and you can think.
The user will specify a task for you to complete. You likely need to run several actions
in order to complete the task. You will only be able to execute a single action at a time.

Use the tools (actions) that are at your disposal. 
Each time you invoke a tool, provide a one-sentence summary of why you are invoking it
and what you expect to accomplish by invoking it.

Your working directory is \texttt{\{LOCAL\_WORKSPACE\_DIR!s\}}.

You will continue the process of invoking tools until you have completed the task.
\end{tcolorbox}

\begin{tcolorbox}[title={\texttt{Install Repository User Instructions}}]
Clone and locally set up the \texttt{{definition.repo.name}} repository from GitHub.
Follow these steps:
1. Git clone the repository \texttt{{definition.repo.info()}} into the directory `\texttt{{install\_path}}`.
2. Check the README (find it if it is not in the root directory) and closely follow the recommended instructions to set up the entire repository correctly for the user.
3. Follow the instructions in the README to correctly set up the repository for the user. Perform any necessary installations, configurations, downloads or setups as described. If the repository is in Python, prefer using `pip` as opposed to conda, virtualenv, or similar. Install the repository and its dependencies globally. Do not use Docker or similar container tools (even if the README suggests it); instead, install the repository and its dependencies globally.
4. Make sure that you complete every step, so that a user could directly use this repository without the need to do further setups, installations or downloads. This includes downloading any necessary pretrained models. However, do NOT download any datasets.
If you encounter any issues, try to solve them.

\texttt{{environment\_variables\_prompt(definition.repo)}}

You should set up the repository in such a way that it can be used to implement the following task later on:
<intended\_task>
<description>
\texttt{{definition.description}}
</description>
<arguments>
\texttt{{"\\n".join(f"{name} ({arg.type}): {arg.description}" for name, arg in definition.arguments.items())}}
</arguments>
<returns>
\texttt{{definition.description\_of\_returns()}}
</returns>
</intended\_task>
IMPORTANT: Your task right now is to only set up the repository, NOT implement this task.

When you are done, provide a brief summary of what you did and what you accomplished, as well as the absolute path to the cloned and installed repository.
\end{tcolorbox}

\begin{tcolorbox}[title={\texttt{Explore Repository User Instructions}}]
\# Background
The repository `\texttt{{definition.repo.name}}` is fully set up and installed at `\texttt{{get\_local\_install\_path(definition.repo)!s}}`.
We need to wrap a specific functionality from this repository into a standalone python function, that can be called independently. 
This function will be called `\texttt{{definition.name}}`, and it is described as follows:
<description>
\texttt{{definition.description}}
</description>

The function will have the following arguments:
<arguments>
\texttt{{"\\n".join(f"<argument>{arg!r}</argument>" for arg in definition.arguments)}}
</arguments>

\texttt{{installed\_repository\_summary}}

\# High-level approach
In order to implement this function, you will follow these steps:
1. Explore the repository to gather all relevant information needed to write the plan.
2. Write a plan for the body/implementation of the function. This plan should be in the form of very high-level pseudo-code, that describes how the function will work.
3. Write the function, based on the plan.

\# Task
Right now, you are at step 1: Explore the repository to gather all relevant information needed to write the plan.
This step is very important, and you must be thorough because you will rely on this information later when implementing the function.
To gather all relevant information needed for the implementation, explore the repository, but only look at relevant files.
Use the tools at your disposal to read files, list directories, search, etc.
HINT 1: If the repository contains a README file, that is often a good starting point. Note that there may be zero or more README files. Always check for README files, and prefer to follow the instructions therein.
HINT 2: If the repository provides a command line interface, prefer to invoke that via subprocess, rather than calling the underlying python functions. Only as a last resort, wrap python functions.
HINT 3: Do NOT attempt to read image files, audio files, etc.
Do not unnecessarily read files that are not relevant for the task.
**However, make sure to read ALL files (e.g. documentation, code, configuration files, etc.) that are necessary in order to implement the function. It should be possible to implement the function based only on the plan and the files you read!**
**You should read relevant code files in order to understand how the functionality you are wrapping is implemented. If you are planning to wrap specific functions, be sure to read the relevant code in order to understand what the input and output arguments/formats are. This is especially relevant if the function you are wrapping produces output files that you will need to read.**
Do NOT write the function yet.
Your task is specifically to explore the repository to gather information.

Once you have gathered ALL relevant information, respond with a one-paragraph summary of what you found.

Remember, the function should do the following:
<description>
\texttt{{definition.description}}
</description>

As such, the signature of the function will be:
```python
\texttt{{definition!s}}
```
\end{tcolorbox}

\begin{tcolorbox}[title={\texttt{Explore Repository User Instructions}}]
Using the information you gathered previously, your task is now to write an outline (plan) for the body/implementation of the function. 
    This plan should be in the form of very high-level pseudo-code, that describes how the function will work.
    It should be a numbered list of steps, each of which describes what you will do in that step.
    Respond with just this list of steps, nothing else.
    Remember, the function should do the following: `{definition.description}`
    
    As such, the signature of the function will be:
    ```python
    \texttt{{definition!s}}
    ```
\end{tcolorbox}

\begin{tcolorbox}[title={\texttt{Summarize Problem User Instructions}}]
Provide a one-paragraph summary of the most recent problem that occured, your diagnosis of it, and how you attempted to fix it with this code change. 
Be specific. 
Include any file paths and other details that are relevant to the problem/solution. 
Your summary should contain all information needed to implement the fix, and include the key insights/observations made for diagnosing the problem.
Begin your response with "The problem was..."
\end{tcolorbox}

\begin{tcolorbox}[title={\texttt{Summarize Problem User Instructions}}]
Now that you have identified the problem as well as a plan to fix the function, you need to write the updated implementation of the function.
Remember, the function is called `{definition.name}`, and it is described as follows: `\texttt{{definition.description}}`.
The function will have the following arguments:
<arguments>
\texttt{{"\\n".join(f"<argument>{arg!r}</argument>" for arg in definition.arguments)}}
</arguments>

As such, the signature of the function will be:
```python
\texttt{{definition!s}}
```

Your task is now to write the Python function.
To do so, use the information you gathered above to fix the function.

\texttt{{coding\_instructions(definition)}}

As a reminder, the current draft of the function is:
```python
\texttt{{code\_draft}}
```

Remember, your diagnosis is:
<diagnosis>
\texttt{{diagnosis.diagnosis}}
</diagnosis>

And your plan to fix the issue is:
<plan>
\texttt{{diagnosis.plan}}
</plan>

Respond with the updated function code only, without any other text.
\end{tcolorbox}

\begin{tcolorbox}[title={\texttt{Coding Instructions}}]
You **must** output a valid, standalone python function that is callable without any modification by a user.
The requirements for the code are:
1. Import the required modules/libraries.
2. You are only allowed to write a single python function. It must start with 'def ...' and end with 'return ...'.
3. You are not allowed to output free texts, test code for the function or anything outside of the function definition.
4. The function needs to be a standalone function that can be called independently.
5. Make sure all required imports are included in the function.
6. The function must perform the task you are given. As a reminder, the task is: `{definition.description}`.
7. Make sure the function accepts all required parameters as inputs.
8. The function must have type hints and a docstring.
9. The function must be named exactly `{definition.name}`.
10. The function must be a valid python function, that can be executed by a python interpreter.

\texttt{{environment\_variables\_prompt(definition.repo)}}

Additional instructions:
* Write the function in such a way that it can easily be debugged later. This means that you should include a lot of print statements for logging purposes. Especially for long-running tasks, it is important to print the progress periodically.
* When catching exceptions in the code (with `try` and `except`), make sure to output the entire stack trace to stderr, so that it can be used to diagnose any issues, e.g. using `traceback.format\_exc()`.
* When running commands and scripts (e.g. using subprocesses), make sure to stream the stdout and stderr to the parent process, so that it can be used to diagnose any issues.
  Use the utility function `run\_and\_stream\_command` provided by the `subprocess\_utils` module. It accepts the same arguments as `subprocess.Popen`, and returns a tuple `(return\_code, output)` (return\_code is an integer, output is a string containing stdout and stderr combined).
  The `run\_and\_stream\_command` automatically handles the streaming of stdout and stderr to the parent process. Be sure to appropriately set the `cwd` argument.
  Example usage:
  ```python
  from subprocess\_utils import run\_and\_stream\_command  \# you must import this
  return\_code, output = run\_and\_stream\_command("echo hello \&\& echo world", shell=True, env={{"MY\_VAR": "my\_value"}}, cwd="/workspace/my\_project")  \# shell=True is the default
  ```
* Make sure that you do not run interactive commands. If some python function that you are calling itself runs interactive commands, try and find a way to avoid calling that function. If, as a last resort, you cannot avoid calling that function, mock/patch the external interactive function to ensure that it does not run interactive commands.
* Always prefer to import existing functions into the function you are writing, or run existing scripts/modules (e.g. via the subprocess functionality descibed above), instead of writing your own implementations. Only if this does not work, or there is no existing function that can be imported, write your own implementation.
\end{tcolorbox}

\begin{tcolorbox}[title={\texttt{Implement Function User Instructions}}]
Now that you have identified the plan for the implementation, you need to write the actual implementation of the function.
This needs to be a standalone python function, that can be called independently. 
This function will be called `\texttt{{definition.name}}`, and it is described as follows: `\texttt{{definition.description}}`.
The function will have the following arguments:
\texttt{{"\\n".join(("- " + repr(arg)) for arg in definition.arguments)}}

As such, the signature of the function will be:
```python
{definition!s}
```

Your task is now to write the Python function.
To do so, follow the plan you identified earlier for the implementation:
<plan>
{plan}
</plan>

\texttt{{coding\_instructions(definition)}}

Remember, you should use the repository `\texttt{{definition.repo.name}}` (installed at `\texttt{{get\_local\_install\_path(definition.repo)!s}}`) to complete the task.
Finally, ensure your function is ready-to-use without any modifications by a user. In many cases, wrapping an existing function, script or module in a subprocess is enough.
Respond with the code of the function only, without any other text.
\end{tcolorbox}

\begin{tcolorbox}[title={\texttt{Diagnose User Instructions}}]

Your initial code implementation did not work. This was attempt number \texttt{{len(problem\_summaries)}} to fix the problem.

Here is a summary of the previous problems, and your attempts to fix them. Keep this in mind as we proceed, and avoid repeating the same mistakes.
<summaries>
\texttt{{'\\n'.join(f'<summary number={i}>{summary}</summary>' for i, summary in enumerate(problem\_summaries))}}
</summaries>

The current version of your code (after \texttt{{len(problem\_summaries)}} attempts) is below.
IMPORTANT: this is the most up-to-date version of your code, so focus on it when diagnosing the problem.
```python
\texttt{{code}}
```

Upon executing this updated function, I received another error.
As a diligent software engineer AI, your task is now to diagnose the issue and fix the function.
You can't see, draw, or interact with a browser, but you can read and write files, and you can run commands, and you can think.
You will be provided with the stdout and stderr from the function execution.
First, use your tools (e.g. running commands, listing directories, reading files, etc.) to gather information about the issue, in order to diagnose it.
Specifically, try to find out the root cause of the issue. Often, this requires reading relevant code files in the repository to understand how the problem occured, and if any assumptions you made in your implementation of the function are incorrect.
Then, formulate a plan to fix the issue, and finally respond with that plan.

NOTE: The plan you write should be the immediate plan to modify the function to fix the issue. 
After you provide the plan, you will then be asked to provide the code to implement the plan, and I will execute that code. 
I will then give you the output of the code execution, and you will be asked to provide a new plan to fix the new issue. 
Therefore, if after exploring the codebase you still don't know what's wrong, your plan should be to modify the function to provide more logging to help you diagnose the problem next time it is executed.

IMPORTANT: While you are able to interact with the environment (writing files, running commands, etc.), any changes you make will be lost when the function is executed again, as the environment will be reset. Therefore, use this opportunity only to gather information about the issue, and not to fix it.
HINT: After gathering information, you may decide to use a slightly different approach to fix the issue -- if this is the case, include this in your plan! 
HINT: Always prefer importing code from the repository, rather than implementing it yourself. The information you gather may contain code that you can import to fix the issue.

Output (stdout and stderr) of the function execution:
<output>
\texttt{{output.stdout}}
</output>

Initial assessment why the function call was not successful:
<assessment>
\texttt{{assessment.reasoning}}
</assessment>

As mentioned above, your immediate task is to diagnose the issue, and formulate a plan to fix it.
\end{tcolorbox}

\begin{tcolorbox}[title={\texttt{Function Execution Assessment User Instructions}}]
I executed the function you wrote.
Based on the output and returned result, assess whether the function call was successful or not.
Specifically, you should assess whether the function performed the task it was supposed to perform.
Also make sure that the returned result is plausible and matches the stdout/stderr output logs, if applicable.
As a reminder, the task is the following:
<task\_description>
\texttt{{definition.description}}
</task\_description>

Description of expected result:
<expected\_result\_description>
\texttt{{definition.description\_of\_returns()}}
</expected\_result\_description>

Returned result:
<result>
\texttt{{truncate\_str(repr(output.result), max\_length=10000)}}
</result>

Output (stdout and stderr) of the function execution:
<output>
\texttt{{truncate\_str(output.stdout, max\_length=10000)}}
</output>

**IMPORTANT: You must also ensure that the returned result itself is correct. This includes ensuring that the result dict contains the correct keys and values, and that the values have the correct types and shapes! If any of these are incorrect, the function call is NOT successful! If this is the case, include this in your reasoning.**
""",
\end{tcolorbox}

\section{OpenHands baseline prompt}
\label{app:openhands_prompts}
Below is the prompt used for the OpenHands baseline~\cite{wang2024openhands}.
\begin{tcolorbox}[title={\texttt{OpenHands Instructions}}]
Your task is to create a tool from the repository {definition.repo.name} which implements the function `{definition.name}` to perform the following task: `{definition.description}`.
While you may perform any necessary installations, configurations, downloads or setups, your deliverables are the following two files:
1. A bash script, named `/workspace/install.sh` that will install all necessary dependencies for the tool to run.
2. A Python file, named `/workspace/code.py` that will contain the code for the tool.

\# Part 1: Install the repository
Clone and locally set up the \texttt{{definition.repo.name}} repository from GitHub.
Follow these steps:
1. Git clone the repository \texttt{{definition.repo.info()}}.
2. Check the README (find it if it is not in the root directory) and closely follow the recommended instructions to set up the entire repository correctly for the user.
3. Follow the instructions in the README to correctly set up the repository for the user. Perform any necessary installations, configurations, downloads or setups as described. If the repository is in Python, prefer using `pip` as opposed to conda, virtualenv, or similar. Install the repository and its dependencies globally.
4. Make sure that you complete every step, so that a user could directly use this repository without the need to do further setups, installations or downloads. This includes downloading any necessary models. However, do NOT download any datasets.
If you encounter any issues, try to solve them.

\texttt{{environment\_variables\_prompt(definition.repo)}}

\# Part 2: Implement the tool function
You need to implement a standalone python function, that can be called independently. 
This function will be called `\texttt{{definition.name}}`, and it is described as follows: `\texttt{{definition.description}}`.
The function will have the following arguments:
\texttt{{"\\n".join((f"- {arg\_name} ({arg.type}): {arg.description}") for arg\_name, arg in definition.arguments.items())}}

As such, the signature of the function will be:
```python
\texttt{{definition!s}}
```
You **must** output a valid, standalone python function that is callable without any modification by a user.
The requirements for the code are:
1. Import the required modules/libraries.
2. You are only allowed to write a single python function. It must start with 'def ...' and end with 'return ...'.
3. You are not allowed to output free texts, test code for the function or anything outside of the function definition.
4. The function needs to be a standalone function that can be called independently.
5. Make sure all required imports are included in the function.
6. The function must perform the task you are given. As a reminder, the task is: `{definition.description}`.
7. Make sure the function accepts all required parameters as inputs.
8. The function must have type hints and a docstring.
9. The function must be named exactly `{definition.name}`.
10. The function must be a valid python function, that can be executed by a python interpreter.

\texttt{{environment\_variables\_prompt(definition.repo)}}

Remember, you should use the repository `{definition.repo.name}` to complete the task.
Finally, ensure your function is ready-to-use without any modifications by a user. In many cases, wrapping an existing function, script or module in a subprocess is enough.
Note: It may be useful to run the function with the following example invocation to test it:
```python3
from code import \texttt{{definition.name}}
\texttt{{definition.name}({", ".join(f"{k}={v!r}" for k, v in definition.example.arguments.items())})}
```

\# IMPORTANT:
- The only two files that you need to produce are `/workspace/install.sh` and `/workspace/code.py` (though you may create other files as well, or install additional dependencies in the process).
- You may use any tools at your disposal to complete the task.
- From within a fresh environment (i.e. a fresh Docker image of python:3.12) that contains the `/workspace` directory which is empty except for your `install.sh` and `code.py` files, it should be possible to run the `install.sh` script, and then run the `code.py` file, without any additional prior installations or dependencies.
- The `code.py` file should NOT contain any imports at the top of the file. The first line of the file should be the function signature (of the `{definition.name}` function). In the body of the function, you may import any necessary modules.
\end{tcolorbox}

\end{document}